\documentclass[runningheads]{llncs}

\usepackage{eccv}

\usepackage{eccvabbrv}

\usepackage{graphicx}
\usepackage{booktabs}

\usepackage[accsupp]{axessibility}  %

\definecolor{linkColor}{rgb}{0.18,0.39,0.62}
\usepackage[breaklinks,colorlinks,citecolor=linkColor]{hyperref}

\usepackage{lipsum}
\usepackage{wrapfig}

\usepackage{orcidlink}

\usepackage{booktabs}
\usepackage{multirow}
\usepackage{colortbl}
\usepackage{soul}

\usepackage{tcolorbox}
\newcommand{\VarSty}[1]{\textnormal{\ttfamily\color{blue!90!black}#1}\unskip}

\definecolor{mygray}{gray}{.92}
\definecolor{mygreen}{rgb}{0.0, 0.6, 0.0}

\usepackage{graphicx} %
\usepackage{tcolorbox} 
\usepackage{multirow}
\usepackage{xcolor,amsmath}
\usepackage[T1]{fontenc}
\usepackage[linesnumbered,ruled,vlined]{algorithm2e}
\SetKwProg{For}{for}{:}{}
\newcommand{\var}{\texttt}

\def\datasetname{AS-V2}
\def\modelname{ASMv2}
\def\benchmarkname{CRPE}

\newcommand\blfootnote[1]{%
\begingroup
\renewcommand\thefootnote{}\footnote{#1}%
\addtocounter{footnote}{-1}%
\endgroup
}

\begin{document}

\title{The All-Seeing Project V2: Towards General Relation Comprehension of the Open World}

\titlerunning{The All-Seeing Project V2}

\author{
Weiyun Wang\inst{2,1},Yiming Ren\inst{3,1},Haowen Luo\inst{3},Tiantong Li\inst{3,1},Chenxiang Yan\inst{3},
Zhe Chen\inst{5,1} \and
Wenhai Wang\inst{4,1} \and
Qingyun Li\inst{6,1} \and
Lewei Lu\inst{7} \and
Xizhou Zhu\inst{3,1,7} \and
\\
Yu Qiao\inst{1} \and
Jifeng Dai\inst{\dagger 3,1}
}

\authorrunning{W. Wang et al.}

\institute{
$^1$OpenGVLab, Shanghai AI Laboratory \quad
$^2$Fudan University \\
$^3$Tsinghua University \quad
$^4$The Chinese University of Hong Kong \\
$^5$Nanjing University \quad
$^6$Harbin Institute of Technology \quad
$^7$SenseTime Research
}

\maketitle
\blfootnote{\noindent$^{\dagger}$Corresponding to Jifeng Dai <daijifeng@tsinghua.edu.cn>.}

\begin{abstract}

We present the All-Seeing Project V2: a new model and dataset designed for understanding object relations in images.
Specifically, we propose the All-Seeing Model V2 ({\modelname}) that integrates the formulation of text generation, object localization, and relation comprehension into a relation conversation (ReC) task.
Leveraging this unified task, our model excels not only in perceiving and recognizing all objects within the image but also in grasping the intricate relation graph between them, diminishing the relation hallucination often encountered by Multi-modal Large Language Models (MLLMs).
To facilitate training and evaluation of MLLMs in relation understanding, we created the first high-quality ReC dataset ({\datasetname}) which is aligned with the format of standard instruction tuning data.
In addition, we design a new benchmark, termed Circular-based Relation Probing Evaluation ({\benchmarkname}) for comprehensively evaluating the relation comprehension capabilities of MLLMs. Notably, our {\modelname} achieves an overall accuracy of 64.50 on this relation-aware benchmark, surpassing the 55.63 of LLaVA-1.5 by a large margin.
We hope that our work can inspire more future research and contribute to the evolution towards artificial general intelligence.
Our project is released at \url{https://github.com/OpenGVLab/all-seeing}.


\end{abstract}

\section{Introduction}

\begin{figure}[t]
\hsize=\textwidth
\setlength{\abovecaptionskip}{1.5mm}
\centering
\begin{subfigure}{0.3\textwidth}
    \centering
    \includegraphics[width=0.9\textwidth]{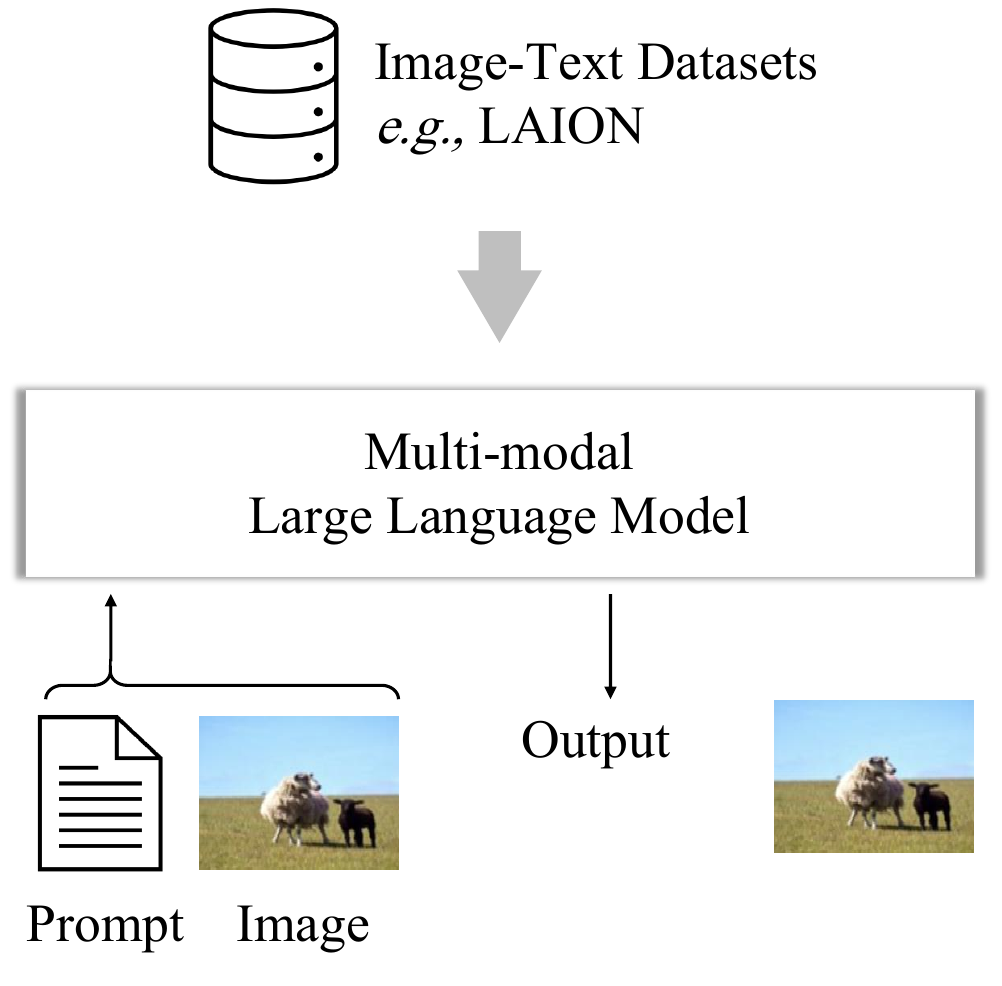}
    \caption{
        Multi-modal Large Language Models (MLLMs) can process both text and images, but they can only capture the holistic visual information of the whole image.
    }
    \label{fig:1a}
\end{subfigure}    
\hspace{0.1in}
\begin{subfigure}{0.3\textwidth}
     \centering
     \includegraphics[width=0.9\textwidth]{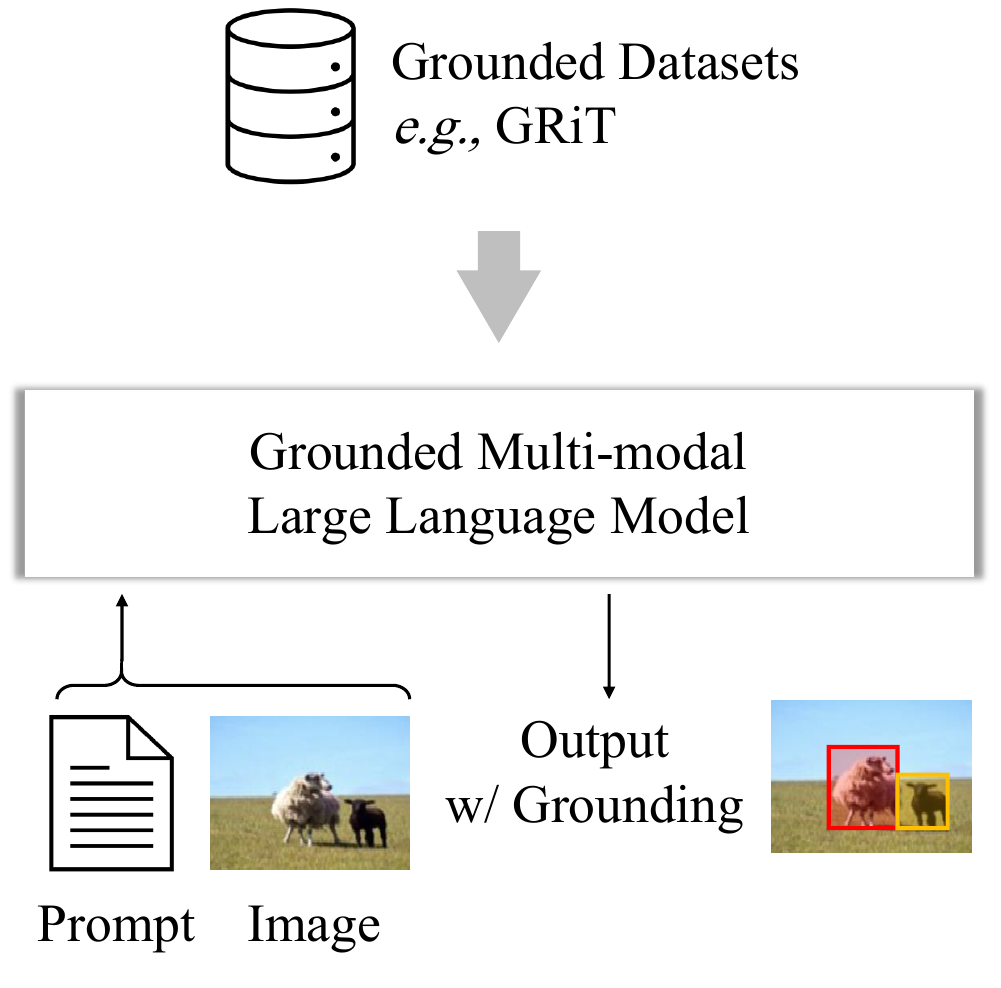}
     \caption{
         Grounded MLLMs can link the objects mentioned in the sentence to the regions in the image while struggling to efficiently understand the relations between objects.
     }
     \label{fig:1b}
\end{subfigure}
\hspace{0.1in}
\begin{subfigure}{0.3\textwidth}
     \centering
     \includegraphics[width=0.9\textwidth]{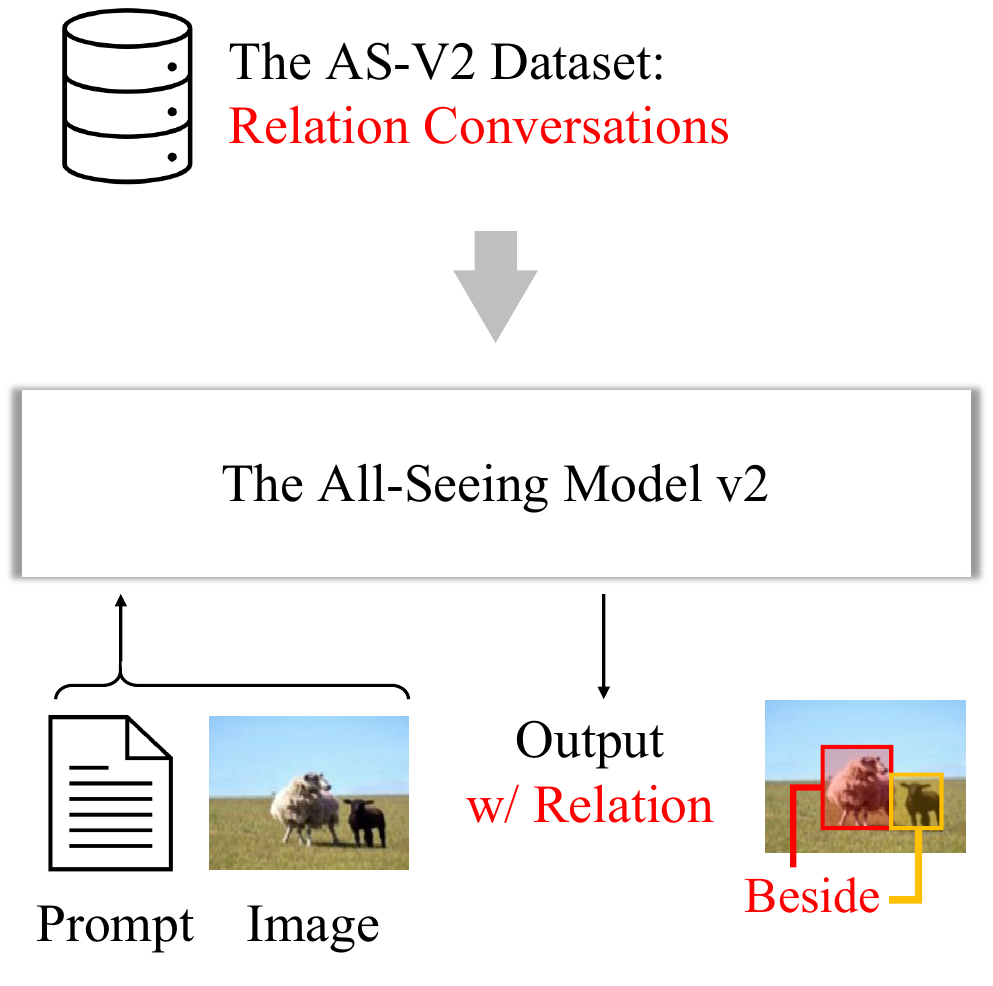}
     \caption{
         Our {\modelname} can comprehend and ground the relations between the objects in the image while maintaining the capabilities of MLLMs and Grounded MLLMs.
     }
     \label{fig:1c}
\end{subfigure}
\caption{
    \textbf{Overview and comparison of our All-Seeing Model v2 with other MLLMs.}
    In this project, we propose
    (1) a novel task, termed Relation Conversation (ReC), which unifies the formulation of text generation, object localization, and relation comprehension;
    (2) a high-quality dataset {\datasetname}, which consists of over 127K samples for ReC;
    (3) the All-Seeing Model v2 ({\modelname}), which is capable of comprehending and grounding the relations between the objects in the image.
}
\label{fig:comparison}
\end{figure}

The study of artificial general intelligence (AGI) systems that can match human intelligence and excel in any task across domains represents the ultimate goal in the field of artificial intelligence.
Benefiting from the advancements of Large Language Models (LLMs), Multi-modal Large Language Models (MLLMs) have demonstrated impressive capabilities in a variety of Vision-Language tasks, suggesting new avenues for achieving AGI.
However, as shown in \cref{fig:1a}, most popular MLLMs~\cite{liu2023llava,liu2023llava_1_5,chen2023internvl} are limited to understanding images as a whole.

As an effective method to improve interaction efficiency, the capabilities of grounding and referring (\ie, adopting bounding boxes in responses) have attracted increasing attention and have been widely integrated into current Grounded MLLMs~\cite{wang2023visionllm,peng2023kosmos2,chen2023shikra,bai2023qwenvl,wang2023allseeing}.
Such capabilities empower models to provide visual responses (\eg, bounding boxes), supporting more vision-language tasks such as region captioning~\cite{mao2016refcoco_plus_g,krishna2017vg}, referring expression comprehension~\cite{kazemzadeh2014refcoco,mao2016refcoco_plus_g}, and referring question answering~\cite{zellers2019vcr}.
However, as shown in \cref{fig:1b}, 
existing models primarily focus on recognizing certain objects within images, overlooking the perception of relations between these objects.
Due to the lack of appropriate modeling methods and suitable training data for relation knowledge, these models struggle to comprehend the inter-object relations within images accurately.
Consequently, these models are prone to hallucinations when dealing with relation questions or overly relying on language priors for judgment.

To enhance relation comprehension ability while maintaining grounding, referring, and other general capabilities, we introduce a novel task, termed Relation Conversation (ReC).
The formulation of ReC unifies the modeling of text generation, object localization, and relation comprehension.
Specifically, as depicted in \cref{fig:1c}, ReC requires the model to generate the text response while linking all mentioned objects as well as the subjects and objects of each predicate in the response to the corresponding regions in the image simultaneously.
Such explicit requirement for predicate grounding challenges the model to comprehend relations between objects within the image.
Notably, models trained on ReC can be naturally adapted to the Scene Graph Generation task.
The grounded objects serve as the nodes in the scene graph while the grounded predicates serve as the edges.
Compared with the traditional scene graph generation, ReC enables the model to generate the scene graph in an open-ended manner, demonstrating the potential to generalize to previously unseen predicate labels, while also maintaining the general ability of MLLMs.

\begin{figure}[!t]
\setlength{\abovecaptionskip}{1.5mm}
\centering
\includegraphics[width=0.95\linewidth]{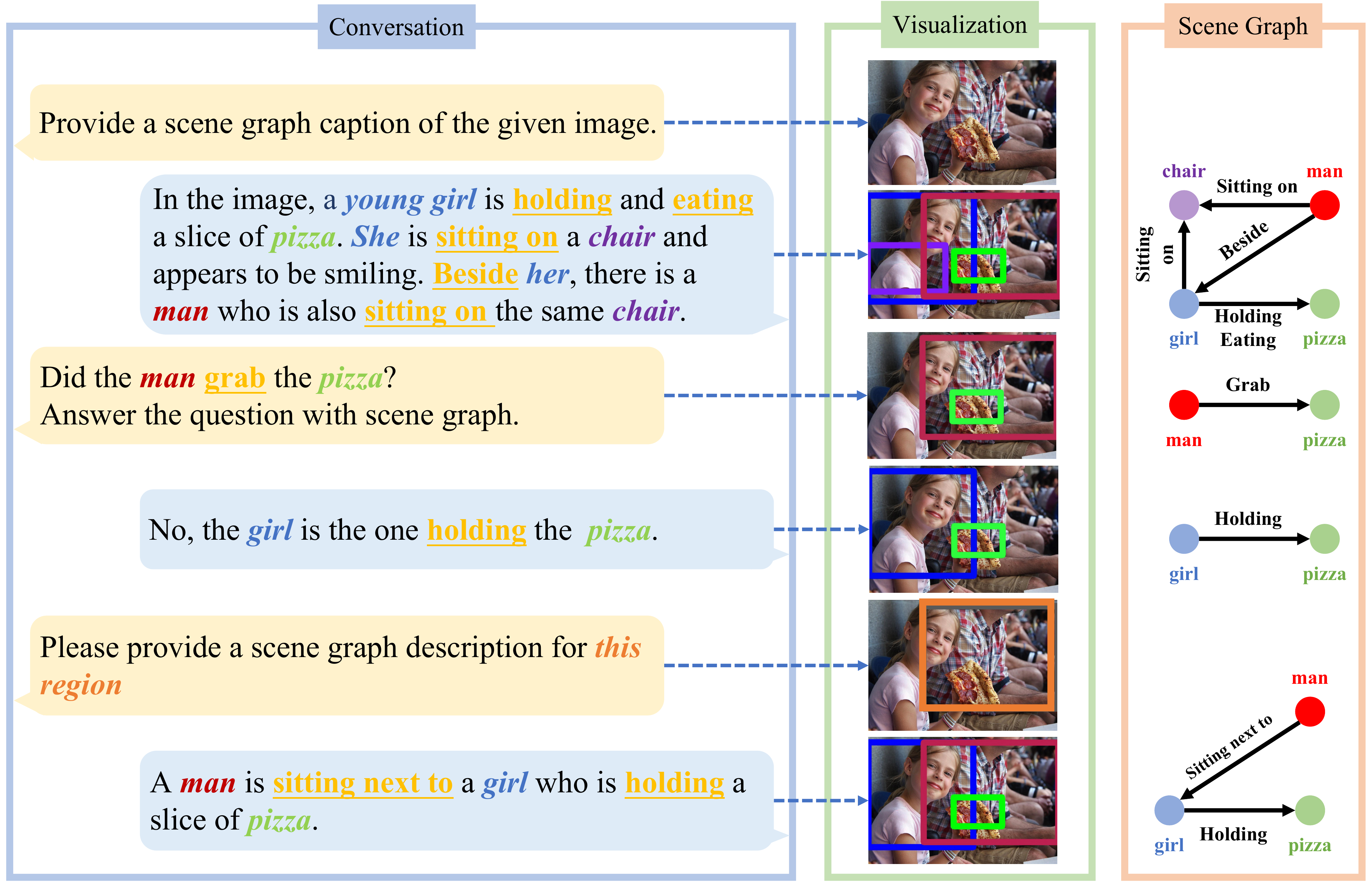}
\caption{
    \textbf{Examples of relation conversation responses from {\modelname}.}
}
\label{fig:qualitative-example}
\end{figure}

From the data aspect,
we construct the All-Seeing Dataset V2 ({\datasetname}) comprising 127K high-quality relation conversation samples, which is built upon the existing caption~\cite{chen2015cococaption}, location~\cite{lin2014microsoft}, and relation~\cite{yang2022psg} annotations.
Combining {\datasetname} with other image-level and region-level multimodal corpora for training, we propose the All-Seeing Model v2 ({\modelname}).
Benefiting from the tailored task format and data, our model can deal with three types of relation tasks, including
(1) Relation Conversation, which requires the model to link all mentioned objects and predicates to the corresponding regions in the image;
(2) Open-ended Scene Graph Generation, which requires the model to generate a scene graph based on the given image in an open-ended manner;
(3) Predicate Classification, which requires the model to generate a scene graph given the ground-truth object labels and localization.
An example of {\modelname} is shown in \cref{fig:qualitative-example}.

To evaluate the relation comprehension ability of existing MLLMs, we construct a benchmark called Circular-based Relation Probing Evaluation ({\benchmarkname}), which is the first benchmark that covers all elements of the relation triplets \texttt{(subject, predicate, object)}, providing a systematic platform for the evaluation of relation comprehension ability.
{\benchmarkname} is formulated as single-choice questions and consists of four splits: Existence, Subject, Predicate, and Object.
The Existence split evaluates the object recognition ability while the remaining splits are designed to evaluate the relation comprehension capability.
Additionally, to evaluate the dependency on language priors, we include abnormal data in {\benchmarkname}, which depict relations that are rare but reasonable in the real world.

Our main contributions are as follows:

(1) We introduce the All-Seeing Project V2, which endows MLLMs with the ability not only to perceive all objects within the image but also to recognize the relations between these objects, leading to superior relation comprehension capability and the potential to generate scene graphs in an open-ended manner.

(2) We propose a novel task, termed Relation Conversation, and the corresponding formulation method, unifying the modeling of captioning, grounding, and relation tasks flexibly.
Based on the task and formulation, we constructed the AS-V2 dataset.
Combining AS-V2 with other general multimodal corpora for training, we propose the All-Seeing Model v2 ({\modelname}), which demonstrates powerful performance across various tasks, including Open-ended Scene Graph Generation and other general image-level and region-level vision-language tasks.

(3) To evaluate the relation comprehension ability of existing MLLMs, we construct the {\benchmarkname} benchmark.
Notably, our {\modelname} achieves an overall accuracy of 64.50 on CRPE, surpassing the 55.63 of LLaVA-1.5 by a large margin.
We also evaluate {\modelname} on various image-level and region-level vision-language tasks.
Specifically, our model achieves an overall score of 74.4 on MMBench~\cite{liu2023mmbench} and 1621.0 on MME~\cite{fu2023mme}, surpassing LLaVA-1.5~\cite{liu2023llava_1_5} by 5.5 points and 90.0 points separately.
Besides, the average accuracy of {\modelname} on grounding benchmarks~\cite{kazemzadeh2014refcoco,mao2016refcoco_plus_g} is 87.42, outperforming Qwen-VL~\cite{bai2023qwenvl} by 1.69 points.

\section{Related Work}

\subsection{Vision-Language Models}
Significant advancements have been made in the field of visual recognition and understanding in recent years.
Models based on the image-text matching framework~\cite{radford2021clip,jia2021align,fang2022eva,chen2023internvl} achieve powerful zero-shot performance on various downstream tasks, thereby initiating the era of open-world semantic recognition and understanding.
Subsequent works~\cite{li2021albef,yu2022coca} further integrate this framework with language modeling tasks to support more generative tasks.
The recent progress of Large Language Models~\cite{brown2020gpt3,openai2023gpt4,touvron2023llama} leads to the emergency of many LLM-based multimodal models~\cite{li2023blip2,zhu2023minigpt4,liu2023llava, wang2023visionllm,li2023videochat,zhai2022lit,chen2023internvl,tian2024mminterleaved,yu2023capsfusion,jiang2024effectiveness,liu2024primitivenet}, which aim to integrate the powerful understanding and reasoning ability of LLMs with multimodal perception and comprehension.
Despite their powerful performance, these works are only capable of capturing the holistic visual information of the whole image.
Some recent methods~\cite{chen2023shikra,peng2023kosmos2,zhang2023gpt4roi,wang2023allseeing,lai2023lisa,zhang2023nextchat,rasheed2023glamm,2023interngpt,gpt4v} begin to focus on location-aware understanding.
However, due to the lack of appropriate modeling methods and training data for relation comprehension, these methods struggle to comprehend the inter-object relations within images accurately.
To enhance relation comprehension ability while maintaining other general capabilities of MLLMs, we introduce a novel task, termed Relation Conversation, which unifies the formulation of text generation, object localization, and relation comprehension.

\subsection{Scene Graph Generation}

Scene Graph Generation (SGG)~\cite{lu2016visual} is a crucial task in scene understanding and has attracted substantial interest across the research community.
This area has witnessed the proposal of diverse model architectures, including message-passing-based frameworks~\cite{li2017vip, dai2017detecting, li2017scene, zellers2018neural, gu2019scene, hu2022neural}, attention-based networks~\cite{zheng2019visual, qi2019attentive}, tree-structured networks~\cite{zhang2017visual, hung2020contextual}, and DETR-based networks~\cite{li2022sgtr, shit2022relationformer, cong2023reltr}.
While most existing methods only utilize images as input, recent works begin to incorporate language information or knowledge graphs to facilitate SGG~\cite{lu2016visual, liao2019natural, zhang2019large, hwang2018tensorize, dupty2020visual,zhou2023vlprompt}, although the scope of language utilization remains limited to basic object or relation concepts.
Compared to prior specialized models, our model is a powerful general model with strong vision-language understanding and reasoning ability and can generate the scene graph in an open-ended manner, exhibiting the potential to generalize to previously unseen predicate labels.

\subsection{Benchmarks for Relation Comprehension}

Evaluating the comprehension of relations between objects is a crucial aspect of advancing MLLMs.
Benchmarks like Visual Genome~\cite{krishna2017vg} and COCO~\cite{lin2014microsoft,chen2015cococaption} provide foundational datasets for object detection and image captioning.
These datasets primarily focus on individual object recognition and general descriptive capabilities. They include annotations for object relations but are not explicitly designed to probe the depth of relation comprehension in a structured and focused manner.
Some synthetic datasets\cite{johnson2017clevr,suhr2017nlvr,andreas2016shapes}, are introduced to probe the spatial reasoning capabilities of vision-language models. These datasets offer controlled environments for model evaluation but inherently limit the problem's scope due to their bounded expressivity.
The Visual Spatial Reasoning (VSR) dataset~\cite{liu2023vsr} asks the model to classify whether the caption correctly describes the relation of two objects presented in the image. This approach primarily focuses on binary classification tasks instead of the understanding of relations within the scene.
In this work, we introduce the {\benchmarkname} benchmark, 
which consists of different splits and each split is designed to probe one of the elements in the relation triplet \texttt{(subject, predicate, object)}.
Therefore, we can evaluate the relation comprehension ability of existing MLLMs more systematically.

\section{Data Construction}

\subsection{The All-Seeing Dataset v2}

\begin{figure}[!t]
\setlength{\abovecaptionskip}{1.5mm}
\centering
\includegraphics[width=\linewidth]{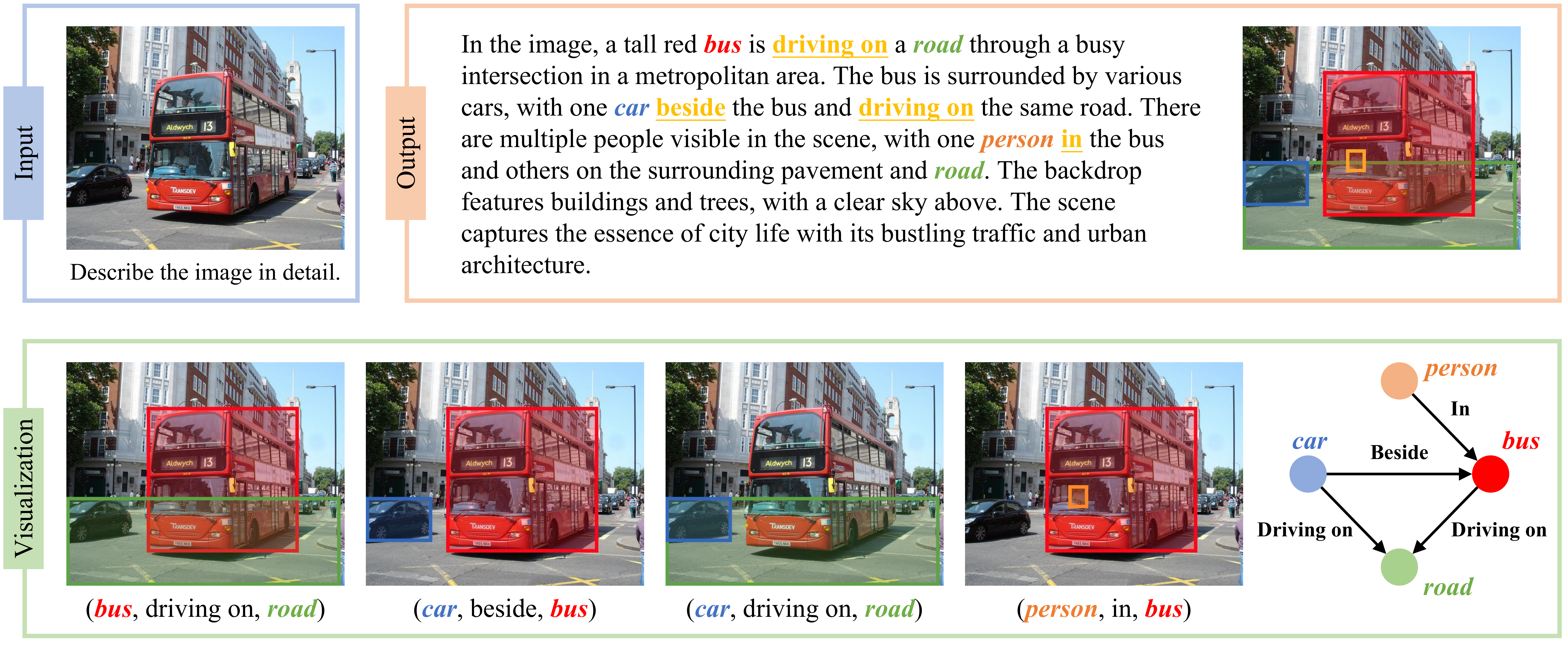}
\caption{
    \textbf{Data example in the {\datasetname} dataset}.
    In the relation conversation, all mentioned objects are linked to their corresponding regions within the image while the predicates are linked to the regions corresponding to their subjects and objects.
}
\label{fig:data-example}
\end{figure}

Our objective is to establish a dataset to unlock the Relation Conversation capability for Multi-modal Large Language Models (MLLMs), which requires the model to predict not only the bounding boxes of each object but also those of the subjects and objects for each predicate mentioned in the sentence.
In this section, we elaborate on the method for constructing the training dataset for ReC, termed All-Seeing Dataset v2 (\datasetname).
Specifically, we utilize GPT-4V~\cite{gpt4v} to construct {\datasetname} based on COCO images~\cite{caesar2018coco} and their annotations~\cite{yang2022psg,lin2014microsoft,chen2015cococaption}.
The key idea is to query GPT-4V to generate responses while linking the objects and predicates mentioned in the generated response to specific regions within the image, referring to the given location annotations and relation annotations.
The formulation of Relation Conversation is presented in \cref{sec:method-sgc}.

The prompt for GPT-4V comprises six components:
(1) \textbf{Task description}, which explains the formulation of relation conversation.
(2) \textbf{Image} to be annotated.
(3) \textbf{Caption annotations} of this image, intended to enhance GPT-4V's understanding of the scene.
(4) \textbf{Location annotations}, which are the bounding boxes of the objects in the scene and guide GPT-4V in annotating the objects in the caption.
(5) {\textbf{Relation annotations}}, which are presented as a list of \texttt{(subject, predicate, object)} triplets and help GPT-4V to annotate the predicate in the caption.
(6) \textbf{Seed examples}, which are manually annotated to assist GPT-4V in comprehending the task description and formatting the output.
Although the caption annotations are not necessary for GPT-4V to produce the desired relation conversation data, incorporating these details into the prompt significantly reduces the hallucinations in the generated data.
An example of the prompt is presented in Appendix \textcolor{red}{B}.

The generated data comprise three types, including:
(1) {\textbf{Detailed description}}, which requires the model to generate a comprehensive description for an image.
(2) {\textbf{Region captioning}}, which requires the model to generate a comprehensive description for a certain region within the image.
(3) {\textbf{Conversation}}, which requires the model to respond to the user query in the multi-turn conversation.
The question types include the relations between objects, the object types, counting the objects, object actions, object locations, and relative positions between objects.
Each type of data is generated using different task descriptions and human-annotated seed examples.
These tasks require the model to understand pointer instructions (\eg, utilizing bounding boxes as prompts) and link the objects and predicates mentioned in the generated response to the image regions.
An example is shown in \cref{fig:data-example}, with more examples in Appendix \textcolor{red}{B}.

In this way, we collected 127K relation conversation samples in total, including 42K in detailed descriptions, 63K in region captioning, and 22K in conversations (90K turns in total), respectively.
The conversation samples also include negative instructions to enhance model robustness. These instances contain incorrect relations, and the model should be able to recognize their incorrectness.

\subsection{Circular-based Relation Probing Evaluation}

\begin{figure}[!t]
\setlength{\abovecaptionskip}{1.5mm}
\centering
\includegraphics[width=\linewidth]{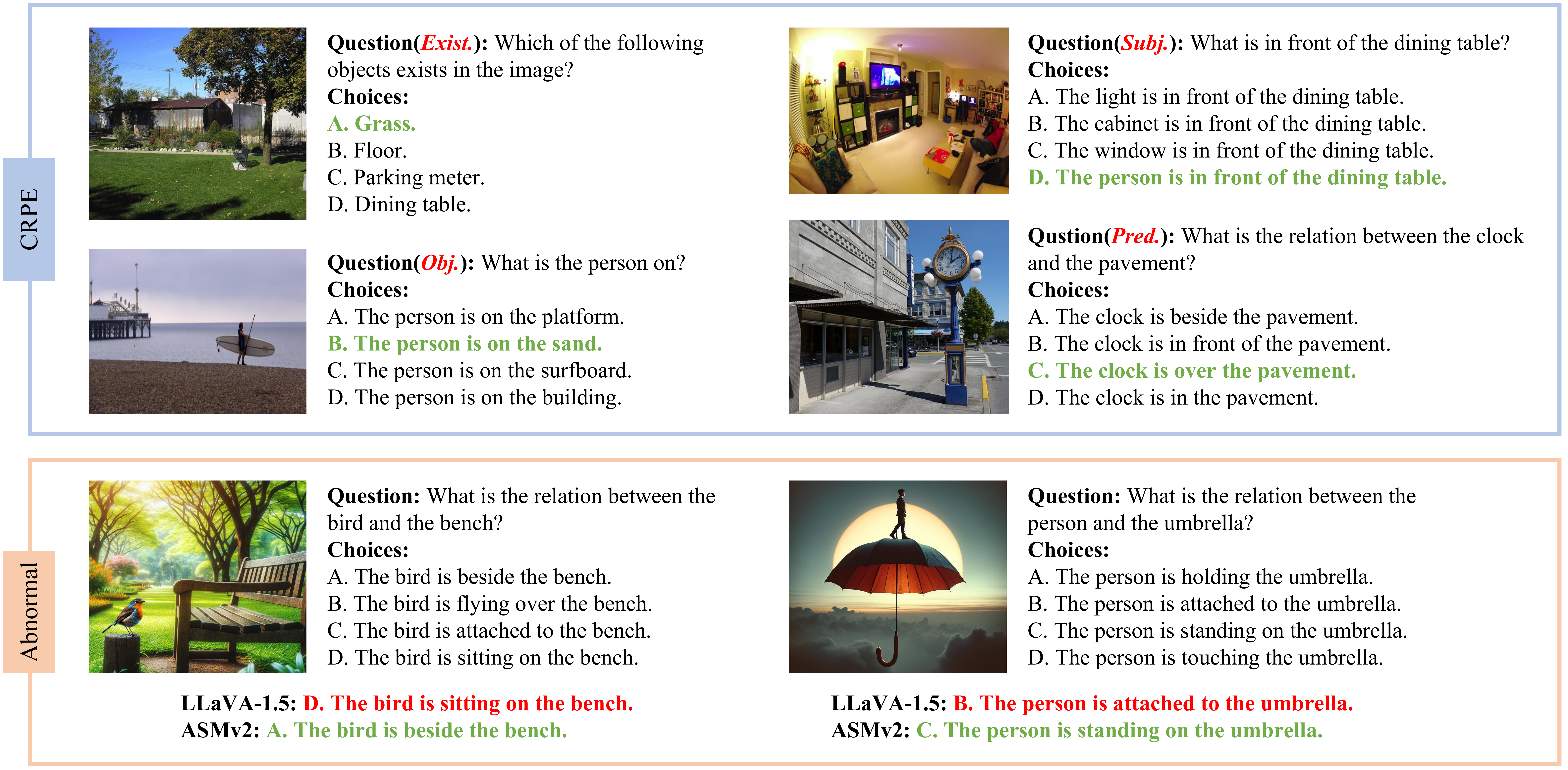}
\caption{
    \textbf{Data examples in the {\benchmarkname}}.
    The benchmark consists of four splits:
    {\textbf{Existence}}, {\textbf{Subject}}, {\textbf{Predicate}}, and {\textbf{Object}}.
    A qualitative comparison in abnormal data between LLaVA-1.5 and {\modelname} is shown at the bottom.
}
\label{fig:benchmark-example}
\end{figure}

In this section, we introduce {\benchmarkname}, a benchmark designed to quantitatively evaluate the object recognition and relation comprehension capabilities of models.
The evaluation is formulated as single-choice questions.
For a robust evaluation, we adopt CircularEval~\cite{liu2023mmbench} as our evaluation strategy.
Under this setting, a question is considered as correctly answered only when the model consistently predicts the correct answer in each of the $N$ iterations, with $N$ corresponding to the number of choices.
In each iteration, a circular shift is applied to both the choices and the answer to form a new query for the model.

As shown in \cref{fig:benchmark-example}, each sample in our benchmark consists of an image and a single-choice question with one correct answer and three wrong answers.
The location annotations~\cite{lin2014microsoft} and the triplets \texttt{(subject, predicate, object)} in the relation annotations~\cite{yang2022psg} are utilized to generate the evaluation data.
We construct four evaluation splits, including
(1) the {\textbf{Existence}} split: the question of this split is ``\textit{Which of the following objects exists in the images?}''. The correct answer is sampled from the semantic tags that exist in the image while the incorrect answer is sampled from those not exist in the image.
(2) the {\textbf{Subject}} split: we generate the question based on the template ``\textit{What is \texttt{<predicate>} the \texttt{<object>}?}'' and consider the subject in the triplet as the correct answer. The negative subjects are sampled from other semantic tags that exist in the image.
(3) the {\textbf{Predicate}} split: we generate the question based on the template ``\textit{What is the relation between \texttt{<subject>} and \texttt{<object>}?}'' and consider the predicate in the triplet as the correct answer. The negative predicates are randomly sampled. Only the predicates satisfying $P(p|s)>0$ and $P(p|o)>0$ can be sampled, where $p,s,o$ refer to predicates, subjects, and objects, separately.
(4) the {\textbf{Object}} split: we generate the question based on the template ``\textit{What is the \texttt{<subject>} \texttt{<predicate>}?}'' and consider the object in the triplet as the correct answer. The negative objects are sampled from other semantic tags that exist in the image.
To avoid reference ambiguity, we ensure that the semantic tags of the subject and object in each triplet are distinct in the image.
We also manually verify the generated samples and filter those with ambiguous questions or choices.

Additionally, to evaluate the dependency on language priors, we further include abnormal data in the Predicate split, which depict relations that are rare but reasonable in the real world.
Specifically, we first select relation triplets with minimal $P(p|s,o)$ and then employ DALLE-3~\cite{betker2023dalle3} to generate corresponding images for these triplets.
Considering that the generated images might not match the specified triplets exactly, we perform a manual filtering process for these triplet-image pairs to ensure data quality.
After that, we generate the evaluation data using the method mentioned above.
\cref{fig:benchmark-example} shows abnormal examples at the bottom and more examples are presented in Appendix \textcolor{red}{D}.

\section{The All-Seeing Model v2}

{\modelname} is a powerful Multi-modal Large Language Model (MLLM), which integrates the Relation Conversation (ReC) ability while maintaining powerful general capabilities. 
Specifically, it follows the model architecture of LLaVA-1.5~\cite{liu2023llava_1_5}, comprising a vision encoder, a vision-language connector, and a language model.
The Vicuna-13B~\cite{zheng2023vicuna} and CLIP-ViT-L-336px~\cite{gao2021clip} are utilized as the language and vision backbone.
This model can deal with three types of relation tasks, including
(1) Relation Conversation, which requires the model to link all objects and predicates mentioned in the response to the corresponding regions in the image.
(2) Open-ended Scene Graph Generation, which requires the model to generate a scene graph based on the given image in an open-ended manner;
(3) Predicate Classification, which requires the model to generate a scene graph given the ground-truth object labels and localization;
In addition, our model is also capable of multi-modality dialogue tasks such as Image Captioning, Visual Question Answering, and Multi-turn conversation.
Since the ReC task requires the model to link the objects and predicates to the corresponding regions in the image, our {\modelname} is also endowed with grounding and referring capabilities and exhibits state-of-the-art performance on region-level tasks.

\subsection{Relation Conversation}
\label{sec:method-sgc}

\begin{figure}[!t]
\setlength{\abovecaptionskip}{1.5mm}
\centering
\includegraphics[width=\linewidth]{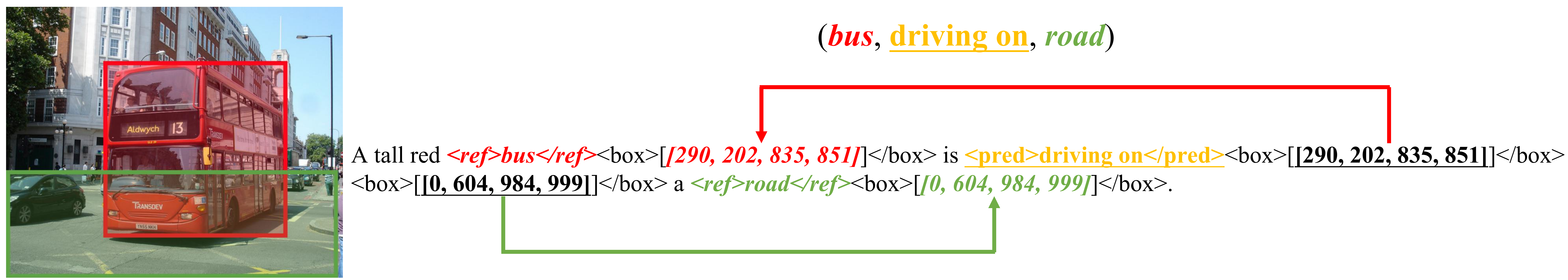}
\caption{
    \textbf{Data formulation for Relation Conversation}.
    Each object is marked with \texttt{<ref></ref>} and followed by a box denoting its location while each predicate is marked with \texttt{<pred></pred>} and followed by two boxes referring to its subjects and objects.
}
\label{fig:data-formulation}
\end{figure}

In this section, we elaborate on the formulation of ReC. Our objective is to propose a task that can enhance the relation comprehension ability while maintaining grounding, referring, and other general capabilities.

As depicted in \cref{fig:data-formulation}, we represent the sentence in the relation conversation as a text sequence. Specifically, our relation conversation marks the object and predicate in the sentence using \texttt{<ref></ref>} and \texttt{<pred></pred>}, respectively.
Each marked object is followed by a bounding box, indicating its localization. Similarly, each predicate is followed by two bounding boxes, which specifically refer to the subjects and objects of the predicate.
All bounding boxes are normalized to integer values within the range [0, 1000) and formatted as:
$\texttt{<box>[[}x_{1}, y_{1}, x_{2}, y_{2}\texttt{]]</box>}$.
Please refer to Appendix \textcolor{red}{A} for more details.

Notably, the response in the relation conversation can be easily parsed into a scene graph.
In a typical scene graph, each node denotes an object in the scene grounded by a bounding box with a semantic label, and each directed edge denotes the relation between a pair of objects with a predicate label.
By utilizing the prediction of bounding boxes for each object (serving as semantic tags for nodes) and those for subjects and objects related to each predicate (serving as nodes, edges, and predicate labels), the generated ReC sentence can be naturally converted into a scene graph.
Nodes without semantic tags will be labeled as Unknown.
To convert the response shown in \cref{fig:data-formulation} into a scene graph,
we first parse the objects marked by ``\texttt{<ref></ref>}'' and assign the marked text as the semantic tag of the following bounding box.
Here, we assign ``\texttt{bus}'' and ``\texttt{road}'' as the semantic tag of the bounding box highlighted in red and green, separately.
Then we extract the predicate label marked by ``\texttt{<pred></pred>}'' (\ie, ``\texttt{driving on}'') and the box coordinates of the subjects and objects of it (\ie, bounding boxes highlighted with \underline{\textbf{bold underline}}).
After that, we utilize these box coordinates as keys to match their semantic tags.
In this example, the bounding box for the subject of ``\texttt{driving on}'' is matched with the box highlighted in red, therefore the subject of ``\texttt{driving on}'' is ``\texttt{bus}''. Similarly, the object of it is ``\texttt{road}''.
Hence, we obtain the parsed triplet \texttt{(bus, driving on, road)}.
Note that the bounding boxes of the subject and object are also part of the triplet while we omit them for simplicity.

Compared with the traditional Scene Graph Generation task, our ReC task exhibits three advantages:
(1) \textbf{More flexible.} Models trained on our proposed Relation Conversation task can be naturally adapted to the Scene Graph Generation task in an open-ended manner.
(2) \textbf{Open-World.} Benefiting from the open-ended generation manner, models trained on ReC have the potential to generalize to previously unseen predicate labels in the Scene Graph Generation task.
(3) \textbf{More general.} ReC requires models to generate a text response and link all mentioned objects to their corresponding regions within the image, thereby maintaining the grounding, referring, and general capabilities of MLLMs and broadening the applicability of these models in real-world scenarios.

\subsection{Model Training}
\label{sec:method-training}
The training process of {\modelname} is divided into two stages, with each stage comprising a pre-training phase and an instruction-tuning phase. 
The first stage is designed to enable the model to effectively understand visual information at the image level.
The pre-training phrase utilizes 595K samples from CC3M~\cite{sharma2018cc3m} filtered by LLaVA~\cite{liu2023llava} while the instruction-tuning phrase utilizes a blend of 665K samples from LLaVA-1.5~\cite{liu2023llava_1_5}.
We update the data format of the region-level data~\cite{kazemzadeh2014refcoco,mao2016refcoco_plus_g,krishna2017vg} in LLaVA-1.5 to the format introduced in \cref{sec:method-sgc}.
The second stage trains the model with a mixture of image-level data and region-level data, which enables the model to comprehend the visual information at the region level, facilitating effective grounding of objects and predicates within sentences.
The pre-training phrase employs 5M samples from CC12M~\cite{sharma2018cc3m} filtered by BLIP~\cite{li2022blip}, 10M filtered samples from AS-1B~\cite{wang2023allseeing}, and 15M filtered samples from GRiT~\cite{peng2023kosmos2}.
The instruction-tuning phase employs 4M samples collected from a variety of sources, including image-level datasets~\cite{liu2023llava,liu2023llava_1_5,zhang2023llavar,chen2023sharegpt4v,liu2023lrv,liu2023vsr,johnson2017clevr,plummer2015flickr30kentities,lu2022sqa,clark2017docqa,biten2019stvqa}, region-level datasets~\cite{kazemzadeh2014refcoco,mao2016refcoco_plus_g,krishna2017vg,wang2023allseeing,zellers2019vcr,zhao2023svit} and our proposed AS-V2 dataset.
The summary of these datasets is presented in Tab. \textcolor{red}{9}.

\section{Experiments}
\label{sec:exp}

In this section, we first compare our {\modelname} with leading Multi-modal Large Language Models (MLLMs) on representative vision-language benchmarks
in \cref{sec:exp-general}.
In addition to these image-level benchmarks, we also evaluate {\modelname} on three representative region-level tasks
in \cref{sec:exp-region}.
After that, {\modelname} is evaluated on the Open-ended Scene Graph Generation task~\cite{yang2022psg} in \cref{sec:exp-sgg}.
The results and analyses of our proposed {\benchmarkname} are presented in \cref{sec:exp-crpe}.
Note that we utilize a consistent checkpoint for all evaluations.

\subsection{Results on General Benchmarks}
\label{sec:exp-general}
\begin{table}[t]
\renewcommand{\arraystretch}{0.85}
\setlength\tabcolsep{0.5mm}
\setlength{\belowcaptionskip}{1.0mm}
\footnotesize
\centering

\caption{
\textbf{Results on 12 general visual-language benchmarks}.
Benchmark names are abbreviated due to space limits. VQA-v2~\cite{goyal2017vqav2}; GQA~\cite{hudson2019gqa}; VizWiz~\cite{gurari2018vizwiz}; SQA$^\text{I}$: ScienceQA-IMG~\cite{lu2022sqa}; VQA$^\text{T}$: TextVQA~\cite{singh2019textvqa}; POPE~\cite{li2023pope}; MME~\cite{fu2023mme}; MMB: MMBench~\cite{liu2023mmbench}; MMB$^\text{CN}$: MMBench-Chinese~\cite{liu2023mmbench}; SEED: SEED-Bench~\cite{li2023seed}; LLaVA$^\text{W}$: LLaVA-Bench (In-the-Wild)~\cite{liu2023llava}; MM-Vet~\cite{yu2023mmvet}. $^*$The training images of the datasets are observed during training.
The best performances are marked \textbf{bold}.
}
\label{tab:general}

\scalebox{0.73}{
\begin{tabular}{@{}l|llclc|ccccccc@{}}
\toprule
Model               & VQA$^\text{v2}$ & GQA & VizWiz & SQA$^\text{I}$ & VQA$^\text{T}$ & POPE & MME & MMB & MMB$^\text{CN}$ & SEED & LLaVA$^\text{W}$ & MM-Vet \\
\midrule
BLIP-2~\cite{li2023blip2}              & 41.0          & 41.0          & 19.6          & 61.0          & 42.5          & 85.3          & 1293.8          & -             & -             & 46.4 & 38.1          & 22.4          \\
InstructBLIP-13B~\cite{instructblip}    & -             & 49.5          & 33.4          & 63.1          & 50.7          & 78.9          & 1212.8          & -             & -             & -    & 58.2          & 25.6          \\
Shikra~\cite{chen2023shikra}              & 77.4*         & -             & -             & -             & -             & -             & -               & 58.8          & -             & -    & -             & -             \\
IDEFICS-9B~\cite{idefics2023}          & 50.9          & 38.4          & 35.5          & -             & 25.9          & -             & -               & 48.2          & 25.2          & -    & -             & -             \\
IDEFICS-80B~\cite{idefics2023}         & 60.0          & 45.2          & 36.0          & -             & 30.9          & -             & -               & 54.5          & 38.1          & -    & -             & -             \\
Qwen-VL~\cite{bai2023qwenvl}             & 78.8*         & 59.3*         & 35.2          & 67.1          & 63.8          & -             & -               & 38.2          & 7.4           & 56.3 & -             & -             \\
Qwen-VL-Chat~\cite{bai2023qwenvl}        & 78.2*         & 57.5*         & 38.9          & 68.2          & 61.5          & -             & 1487.5          & 60.6          & 56.7          & 58.2 & -             & -             \\
LLaVA-1.5-13B~\cite{liu2023llava_1_5}       & 80.0*         & 63.3*         & 53.6          & 71.6          & 61.3          & 85.9          & 1531.3          & 67.7          & 63.6          & 61.6 & 70.7          & 35.4          \\
VILA-13B~\cite{lin2023vila}            & 80.8*         & 63.3*         & \textbf{60.6} & 73.7          & \textbf{66.6} & 84.2          & 1570.1          & 70.3          & 64.3          & 62.8 & 73.0          & 38.8          \\
\rowcolor{mygray}
ASMv2-13B (ours)        & \textbf{81.0}* & \textbf{63.9}* & 58.1          & \textbf{87.1}* & 60.2          & \textbf{86.3} & \textbf{1621.0} & \textbf{74.4} & \textbf{64.3} & \textbf{66.3}    & \textbf{78.9} & \textbf{41.3} \\ \bottomrule
\end{tabular}
}
\end{table}

To evaluate the general ability of {\modelname}, we perform a comprehensive comparison with leading MLLMs in \cref{tab:general}.
Benefiting from the stronger relation comprehension ability, {\modelname} exhibits SoTA performance on these benchmarks.

\subsubsection{Results of Visual Question Answering.}
On general VQA benchmarks, such as VQAv2~\cite{goyal2017vqav2} and GQA~\cite{hudson2019gqa}, our {\modelname} demonstrates superior overall performance compared to LLaVA-1.5~\cite{liu2023llava_1_5} and VILA~\cite{lin2023vila}. On the VQAv2 dataset, our {\modelname} outperforms the LLaVA-1.5-13B by 1.0 points.
Besides, our model also achieves competitive performance with baselines on text-oriented VQA benchmarks, including VizWiz-VQA~\cite{gurari2018vizwiz} and TextVQA~\cite{singh2019textvqa}.

\subsubsection{Results of Multi-modal benchmarks.}
In recent comprehensive benchmarks, which consist of a wide range of sub-tasks covering various fine-grained capabilities, our model significantly outperforms the current SoTA MLLMs, such as LLaVA-1.5~\cite{liu2023llava_1_5} and VILA~\cite{lin2023vila}.
Specifically, our model achieves an overall score of 74.4 on MMBench and 1621.0 on MME, surpassing VILA by 4.1 points and 50.9 points separately.
Besides, {\modelname} also exhibits state-of-the-art performance on SEED~\cite{li2023seed}, LLaVA-Bench~\cite{liu2023llava}, and MM-Vet~\cite{yu2023mmvet}, outperforming baselines by a large margin.
These results demonstrate the general ability of our model.

\subsection{Results on Region-level Benchmarks}
\label{sec:exp-region}

To evaluate the region comprehension and grounding capability, we evaluate {\modelname} on three representative region-level tasks, including (1) Referring Expression Comprehension~\cite{kazemzadeh2014refcoco,mao2016refcoco_plus_g}, which requires the model to localize the target object conditioned on the given description. (2) Region Captioning~\cite{krishna2017vg,mao2016refcoco_plus_g}, which requires the model to generate a caption for a certain object in the image conditioned on the given region. (3) Referring Question Answering~\cite{zellers2019vcr}, which contains region referring in both questions and answers.

\begin{table}[t]
\renewcommand{\arraystretch}{0.85}
\setlength\tabcolsep{0.5mm}
\setlength{\belowcaptionskip}{1.5mm}
\footnotesize
\centering

\caption{
\textbf{Accuracy scores on the Referring Expression Comprehension task}.
}
\label{tab:rec}

\begin{tabular}{@{}l|cccccccc|c@{}}
\toprule
\multirow{2}{*}{Model} & \multicolumn{3}{c}{RefCOCO}                      & \multicolumn{3}{c}{RefCOCO+}                     & \multicolumn{2}{c|}{RefCOCOg}   & \multirow{2}{*}{Avg.} \\
                       & Val            & Test-A         & Test-B         & Val            & Test-A         & Test-B         & Val            & Test           &                       \\ \midrule
OFA-L~\cite{wang2022ofa}                  & 79.96          & 83.67          & 76.39          & 68.29          & 76.00          & 61.75          & 67.57          & 67.50          & 72.64                 \\
VisionLLM-H~\cite{wang2023visionllm}            & -              & 86.70          & -              & -              & -              & -              & -              & -              & -                     \\
Shikra-7B~\cite{chen2023shikra}              & 87.01          & 90.61          & 80.24          & 81.60          & 87.36          & 72.12          & 82.27          & 82.19          & 82.93                 \\
Shikra-13B~\cite{chen2023shikra}             & 87.83          & 91.11          & 81.81          & 82.89          & 87.79          & 74.41          & 84.64          & 83.16          & 84.21                 \\
Qwen-VL-7B~\cite{bai2023qwenvl}        & 88.55          & 92.27          & 84.51          & 82.82          & 88.59          & 76.79          & 85.96          & 86.32          & 85.73                 \\
MiniGPT-V2-7B~\cite{chen2023minigptv2}     & 88.06          & 91.29          & 84.30          & 79.58          & 85.52          & 73.32          & 84.19          & 84.31          & 83.82                 \\
Ferret-13B~\cite{you2023ferret}             & 89.48          & 92.41          & 84.36          & 82.81          & 88.14          & 75.17          & 85.83          & 86.34          & 85.57                 \\
\rowcolor{mygray}
ASMv2-13B (ours)           & \textbf{90.56} & \textbf{94.24} & \textbf{86.24} & \textbf{84.81} & \textbf{90.83} & \textbf{76.89}    & \textbf{87.52} & \textbf{88.26} & \textbf{87.42}        \\ \bottomrule
\end{tabular}
\end{table}

\subsubsection{Results of Referring Expression Comprehension.}
Our {\modelname} achieves state-of-the-art performance on the representative REC benchmarks~\cite{kazemzadeh2014refcoco,mao2016refcoco_plus_g}.
As shown in \cref{tab:rec}, our {\modelname} significantly outperforms current state-of-the-art MLLMs, including Qwen-VL~\cite{bai2023qwenvl} and Ferret~\cite{you2023ferret}.

\subsubsection{Results of Region Captioning.}
\begin{table}[t]
\renewcommand{\arraystretch}{0.85}
\setlength\tabcolsep{0.5mm}
\setlength{\belowcaptionskip}{1.5mm}
\footnotesize

\centering
\caption{
\textbf{Results on the Region Captioning task}.
We mark the best performance \textbf{bold} and the second-best \underline{underlined}.
}
\label{tab:region_captioning}

\begin{tabular}{@{}lcccc@{}}
\toprule
\multirow{2}{*}{Model} & \multicolumn{2}{c}{VG~\cite{krishna2017vg}}         & \multicolumn{2}{c}{RefCOCOg~\cite{mao2016refcoco_plus_g}} \\ \cmidrule(l){2-5} 
                       & METEOR        & CIDEr          & METEOR          & CIDEr            \\ \midrule
GRiT~\cite{wu2022grit}                   & 17.1          & 142.0          & 15.2            & 71.6             \\
SLR~\cite{yu2017slr}                    & -             & -              & 15.4            & 59.2             \\
SLR+Rerank~\cite{yu2017slr}             & -             & -              & 15.9            & 66.2             \\
Kosmos-2~\cite{peng2023kosmos2}       & -             & -              & 14.1            & 62.3             \\
GPT4RoI-7B~\cite{zhang2023gpt4roi}             & 17.4          & 145.2          & -               & -                \\
GPT4RoI-13B~\cite{zhang2023gpt4roi}            & 17.6          & 146.8          & -               & -                \\
ASM-FT~\cite{wang2023allseeing}                 & \textbf{18.3}          & \underline{148.7}          & \textbf{21.8}            & \underline{107.8}            \\
\rowcolor{mygray}
ASMv2-13B (ours)           & \underline{17.9} & \textbf{153.5} & \underline{21.7}   & \textbf{114.7}   \\ \bottomrule
\end{tabular}
\end{table}

Our model demonstrates state-of-the-art performance on the representative region captioning benchmarks, including VG~\cite{krishna2017vg} and RefCOCOg~\cite{mao2016refcoco_plus_g}.
As shown in \cref{tab:region_captioning}, our model achieves a CIDEr score of 114.7 on RefCOCOg, which surpasses the current state-of-the-art model (\ie ASM-FT) by 6.9 points. On the VG dataset, our model also exhibits competitive results compared to the current state-of-the-art model. 

\subsubsection{Results of Referring Question Answering.}
\begin{table}[t]
\renewcommand{\arraystretch}{0.85}
\setlength\tabcolsep{0.5mm}
\setlength{\belowcaptionskip}{1.5mm}
\footnotesize
\centering

\caption{
\textbf{Results on Visual Commonsense Reasoning}.
Q, A, and R denote the \textbf{Q}uestion, \textbf{A}nswer, and \textbf{R}ationale. X$\rightarrow$Y means that the model needs to select the correct option for Y conditioned on X. $^*$The single-task fine-tuning setting.
}
\label{tab:vcr}

\begin{tabular}{@{}lccc@{}}
\toprule
\multirow{2}{*}{Method} & \multicolumn{3}{c}{Validation Acc. (\%)}                          \\ \cmidrule(l){2-4} 
                        & Q$\rightarrow${}A & QA$\rightarrow${}R & Q$\rightarrow${}AR \\ \midrule
ViLBERT~\cite{lu2019vilbert}                 & 72.4              & 74.5               & 54.0               \\
Unicoder-VL~\cite{li2020unicoder}             & 72.6              & 74.5               & 54.5               \\
VLBERT~\cite{su2019vl_bert}                  & 75.5              & 77.9               & 58.9               \\
ERNIE-ViL-L~\cite{yu2021ernie}             & 78.5              & 83.4               & 65.8               \\
VILLA~\cite{gan2020villa}                   & 78.5              & 82.6               & 65.2               \\
*GPT4RoI-7B~\cite{zhang2023gpt4roi}             & 87.4              & 89.6               & 78.6               \\
\rowcolor{mygray}
ASMv2-13B (ours)            & 87.8              & 88.8               & 78.4               \\
\rowcolor{mygray}
*ASMv2-13B (ours)           & \textbf{88.4}     & \textbf{89.9}      & \textbf{79.4}      \\ \bottomrule
\end{tabular}
\end{table}

We evaluate the Referring Question Answering (RQA) ability of {\modelname} on the Visual Commonsense Reasoning (VCR) dataset~\cite{zellers2019vcr}, which evaluates the commonsense reasoning abilities in the form of single-choice questions.
The questions and candidate choices in VCR contain region referring.
The results are presented in \cref{tab:vcr}. Although trained in a multi-task setting, {\modelname} exhibits competitive performance compared to the current state-of-the-art model (\ie, GPT4RoI~\cite{zhang2023gpt4roi}), which is finetuned on the VCR dataset in a single task setting. In addition, after the single task finetuning, our model outperforms GPT4RoI by 0.8 points.

\subsection{Results on Open-ended Scene Graph Generation}
\label{sec:exp-sgg}

In this section, we evaluate the Relation Conversation capability of our model through the Open-ended Scene Graph Generation task on the Panoptic Scene Graph (PSG) dataset~\cite{yang2022psg}, which is a widely-used benchmark for the scene graph generation.
See Appendix \textcolor{red}{C.2} for the results on the Predicate Classification task.

\subsubsection{Baselines.}
Despite the powerful performance, most previous methods~\cite{yang2022psg,zhong2021sggnls,wang2023pairnet,zhou2023vlprompt} are constrained by pre-defined label sets and struggle to capture a diverse range of visual concepts from natural language in an open-ended manner.
On the other hand, TextPSG~\cite{zhao2023textpsg} explores a methodology for generating scene graphs in an open-ended manner, which first generates the region proposals and then asks BLIP~\cite{li2022blip} to predict the semantic tags and predicate labels for these regions auto-regressively.
Here, we consider traditional close-set scene graph generation models and TextPSG as our baseline in OpenSGG.

\subsubsection{Metrics.}
Following the common practice~\cite{yang2022psg,zhao2023textpsg}, we report the triplet Recall and mean Recall for every predicate category (mRecall) in the OpenSGG task.
Concretely, a scene graph consists of a set of triplets \texttt{(subject, predicate, object)}. A triplet is considered to be correct if the phrase labels are all correct and the location of the subject and object should match the ground truth with IoU greater than 0.5 respectively.
We also report {\#Tuples} to denote the average number of predicted tuples for each generated scene graph.

\subsubsection{Results.}

\begin{wraptable}{r}{0.49\textwidth}
\renewcommand{\arraystretch}{0.85}
\setlength\tabcolsep{0.5mm}
\setlength{\abovecaptionskip}{-8mm}
\footnotesize
\centering

\caption{
\textbf{Recall scores on PSG}.
\textcolor{gray!45}{Gray} denotes that the model generates the scene graphs in a close-ended manner.
}
\label{tab:psg}

\begin{tabular}{@{}lrrr@{}}
\toprule
Model        & \#Tuples & Recall & mRecall \\ \midrule
\textcolor{gray!45}{IMP~\cite{xu2017IMP}}    & \textcolor{gray!45}{20.0}     & \textcolor{gray!45}{16.5}   & \textcolor{gray!45}{6.5}    \\
\textcolor{gray!45}{MOTIFS~\cite{zellers2018MOTIFS}}    & \textcolor{gray!45}{20.0}     & \textcolor{gray!45}{20.0}   & \textcolor{gray!45}{9.1}    \\
\textcolor{gray!45}{VCTree~\cite{tang2019vctree}}    & \textcolor{gray!45}{20.0}     & \textcolor{gray!45}{20.6}   & \textcolor{gray!45}{9.7}    \\
\textcolor{gray!45}{GPSNet~\cite{lin2020gpsnet}}    & \textcolor{gray!45}{20.0}     & \textcolor{gray!45}{17.8}   & \textcolor{gray!45}{7.0}    \\
\textcolor{gray!45}{PSGFormer~\cite{yang2022psg}}    & \textcolor{gray!45}{20.0}     & \textcolor{gray!45}{18.6}   & \textcolor{gray!45}{16.7}    \\
\midrule
TextPSG~\cite{zhao2023textpsg}      & 50.0     & 4.8    & -       \\
TextPSG~\cite{yang2022psg}      & 100.0    & 5.5    & -       \\
\rowcolor{mygray}
ASMv2 (ours) & 9.2      & 14.2   & 10.3    \\ \bottomrule
\end{tabular}

\end{wraptable}

As shown in \cref{tab:psg}, our {\modelname} demonstrates state-of-the-art performance in the OpenSGG task.
Specifically, our {\modelname} significantly outperforms TextPSG by 8.7 points in recall while generating a significantly fewer average number of tuples compared to it (9.2 \vs 100.0).
Note that having more tuples generally implies an advantage in computing recall.
When compared to traditional scene graph generation models, which generate scene graphs in a close-ended manner, {\modelname} also exhibits competitive performance.
Despite generating fewer tuples, our model maintains a competitive recall of 14.2 and a mean recall of 10.3.
Another factor negatively impacting the performance is that our {\modelname} generates scene graphs in an open-ended manner while recall is calculated in an exact-match manner.
Therefore, the triplets \texttt{(people, standing on, grass)} and \texttt{(person, standing on, grass)} are considered mismatched even though they represent the same semantics.
A more appropriate metric for this task will be left for future work.

\subsection{Results on {\benchmarkname}}
\label{sec:exp-crpe}

\begin{wraptable}{r}{0.55\textwidth}
\renewcommand{\arraystretch}{0.85}
\setlength\tabcolsep{0.5mm}
\setlength{\abovecaptionskip}{-8mm}
\scriptsize
\centering

\caption{
\textbf{Accuracy scores on {\benchmarkname}}.
}
\label{tab:crpe}

\begin{tabular}{@{}lccccc@{}}
\toprule
\textbf{Model} & \textbf{Exist.} & \textbf{Subj.}  & \textbf{Pred.}  & \textbf{Obj.}   & \textbf{Overall} \\ \midrule
Qwen-VL~\cite{bai2023qwenvl}                       & 85.11              & 45.66          & 38.19          & 31.60          & 38.48            \\
LLaVA-1.5~\cite{liu2023llava_1_5}                          & 88.69              & 57.44          & 54.24          & 55.21          & 55.63            \\
\rowcolor{mygray}
ASMv2 (ours)                       & \textbf{92.14}     & \textbf{69.21} & \textbf{58.95} & \textbf{65.34} & \textbf{64.50}   \\ \bottomrule
\end{tabular}

\end{wraptable}

In this section, we evaluate the relation comprehension ability of our {\modelname} and current leading MLLMs~\cite{liu2023llava_1_5,bai2023qwenvl} using our proposed {\benchmarkname} benchmark.
This benchmark consists of four splits: Existence, Subject, Predicate, and Object.
The Existence split evaluates the models' object recognition ability while the remaining splits are designed to evaluate the models' relation comprehension ability.
In addition to reporting the performance of each split in the benchmark individually, we also report the average score of the latter three splits
as the overall score for relation comprehension ability.

As shown in \cref{tab:crpe}, the performance of existing MLLMs on the Existence questions is significantly higher than on the Subject, Predicate, and Object questions. This suggests that these models have a more robust capability to recognize objects within an image than to comprehend the relations between them.
Specifically, our {\modelname} shows a remarkable improvement in understanding object relations compared to the other models. For example, {\modelname} achieves an overall accuracy of 64.50, which is significantly higher than the 55.63 of LLaVA-1.5 and the 38.48 of Qwen-VL.
These results demonstrate that our model can comprehend the relations between the objects within the image better, benefiting from the training of relation conversation data.

\section{Conclusion}

In this paper, we propose a novel task, termed Relation Conversation (ReC), to challenge the model to understand the relations between the objects within the image.
We construct the All-Seeing Dataset V2 ({\datasetname}), which is a high-quality ReC dataset to unlock the ReC ability of Multi-modal Large Language Models (MLLMs) and the {\benchmarkname} to quantitatively evaluate the relation comprehension ability.
Leveraging {\datasetname} and other general multimodal corpora for training, we introduce the All-Seeing Model v2 ({\modelname}), which exhibits stronger relation comprehension ability compared to existing leading MLLMs and achieves state-of-the-art performance on the Open-ended Scene Graph Generation task and various general image-level and region-level tasks.
We hope that our work can inspire more future research and contribute to the evolution towards artificial general intelligence, equipping artificial intelligence systems with an ``all-seeing eye'' to achieve a deeper understanding of the world.

\section*{Acknowledgments}

The work is supported by the National Key R\&D Program of China (NO. 2022ZD0161300), and the National Natural Science Foundation of China (Grant No. 62376134).

\bibliographystyle{splncs04}
\bibliography{egbib}

\clearpage
\appendix

\begin{figure}[!h]
\vspace{-5mm}
\centering
\includegraphics[width=\linewidth]{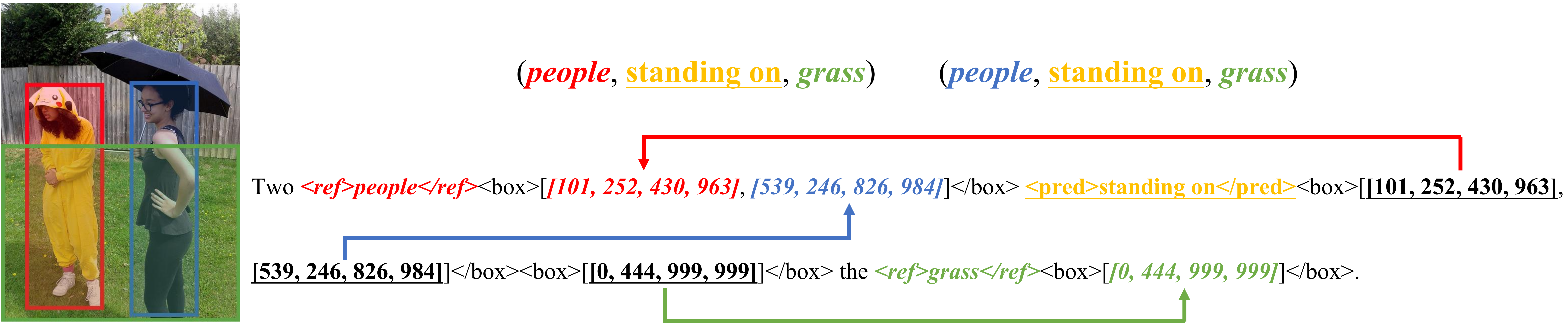}
\caption{
    \textbf{A complex example for the formulation of Relation Conversation}.
}
\label{fig:data-formulation-complex}
\vspace{-5mm}
\end{figure}

\section{Relation Conversation}
\label{sec:appendix-method-rec}

In this section, we introduce more details about the formulation of Relation Conversation.
As depicted in \cref{fig:data-formulation-complex}, when a certain text span is associated with multiple regions, these bounding boxes are formatted as:

\begin{equation*}
    \texttt{<box>[[}x_{1}^{1}, y_{1}^{1}, x_{2}^{1}, y_{2}^{1}\texttt{], ..., [}x_{1}^{n}, y_{1}^{n}, x_{2}^{n}, y_{2}^{n}\texttt{]]</box>},
\end{equation*}

\noindent
where $\texttt{[}x_{1}^{i}, y_{1}^{i}, x_{2}^{i}, y_{2}^{i}\texttt{]}$ denotes the $i$-th bounding box linked to the object or predicate.
For a specific predicate, the subject and object must be linked to an equal number of bounding boxes. Otherwise, one of them must be linked to just one bounding box and thus can be broadcast to match the count of another one.

As shown in \cref{fig:data-formulation-complex},
to parse this example into a scene graph, we first assign the semantic tag ``\texttt{people}'' to the bounding box highlighted in red and blue. Similarly, we assign the semantic tag ``\texttt{grass}'' to the bounding box highlighted in green.
We then extract the predicate label enclosed in ``\texttt{<pred></pred>}'' (\ie, ``\texttt{standing on}'') and the box coordinates of its subjects and objects (\ie, bounding boxes highlighted with \underline{\textbf{bold underline}}).
After that, we utilize these box coordinates as keys to match their respective semantic tags.
Considering that two subjects are linked to the predicate while only one object is linked, we broadcast the object to match the number of subjects.
Given $N$ subjects and $N$ objects, we pack them into $N$ tuples in order, where each tuple consists of one subject and one object. In this example, we obtain two tuples, resulting in two parsed triplet \texttt{(people, standing on, grass)}, each connected to a \texttt{people} with different bounding boxes.

\section{The All-Seeing Dataset v2}
\label{sec:appendix_asv2}
\label{sec:appendix_asv2_prompts}

\begin{figure}[h!]
\hsize=\linewidth
\centering
\begin{subfigure}{0.95\linewidth}
    \centering
    \includegraphics[width=\linewidth]{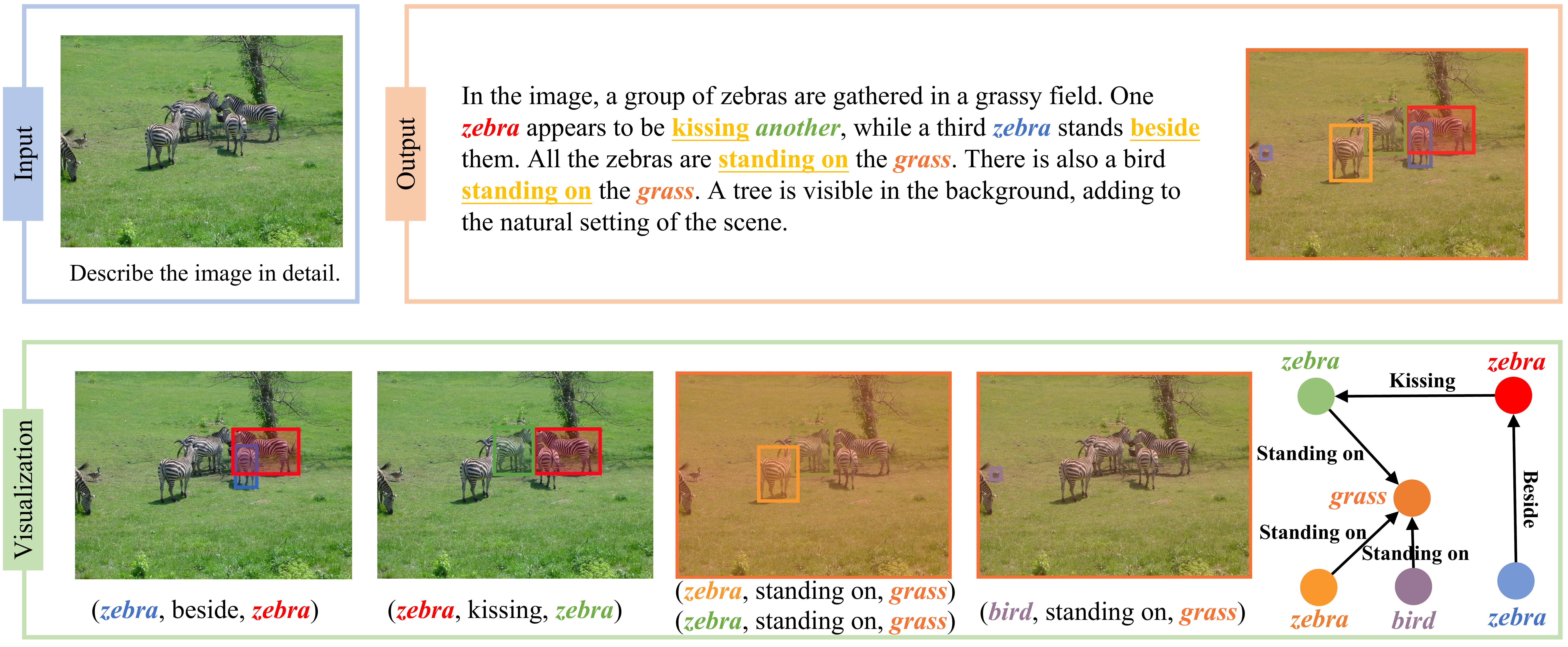}
    \caption{}
\end{subfigure}    
\begin{subfigure}{0.95\linewidth}
     \centering
     \includegraphics[width=\linewidth]{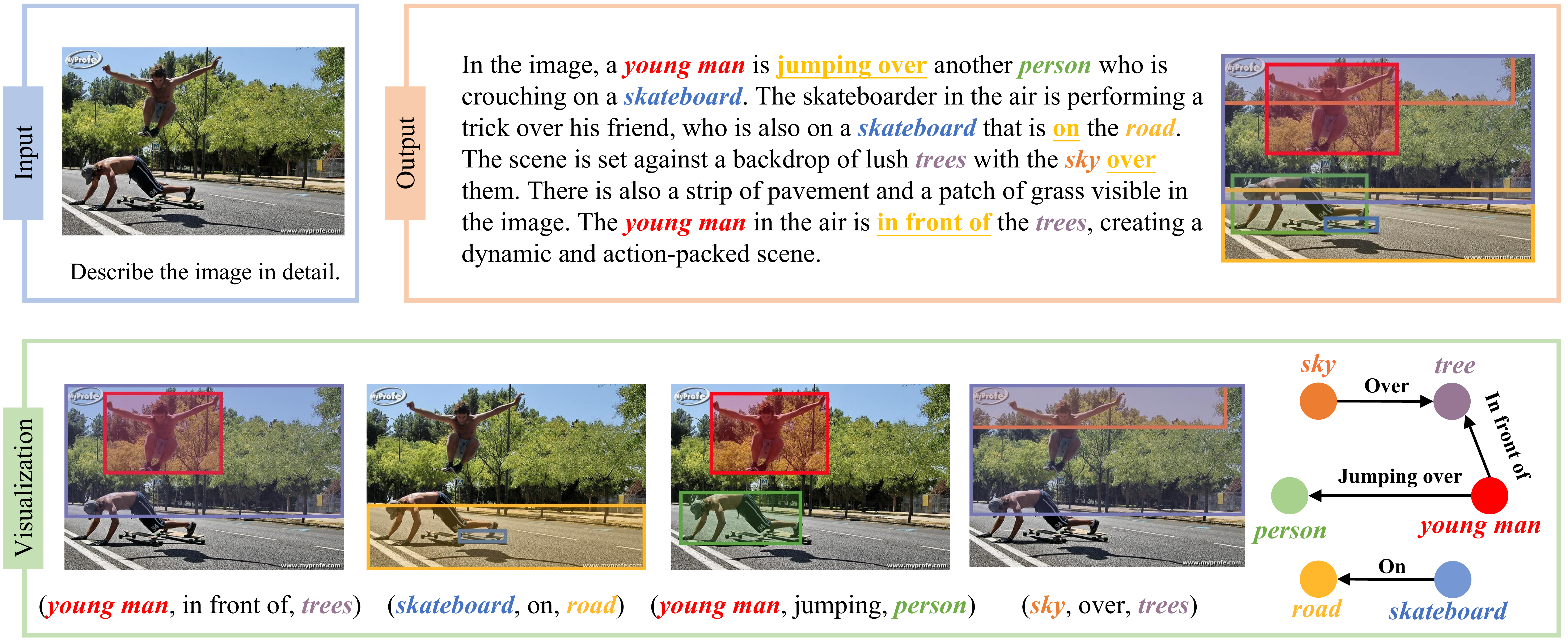}
     \caption{}
\end{subfigure}
\begin{subfigure}{0.95\linewidth}
     \centering
     \includegraphics[width=\linewidth]{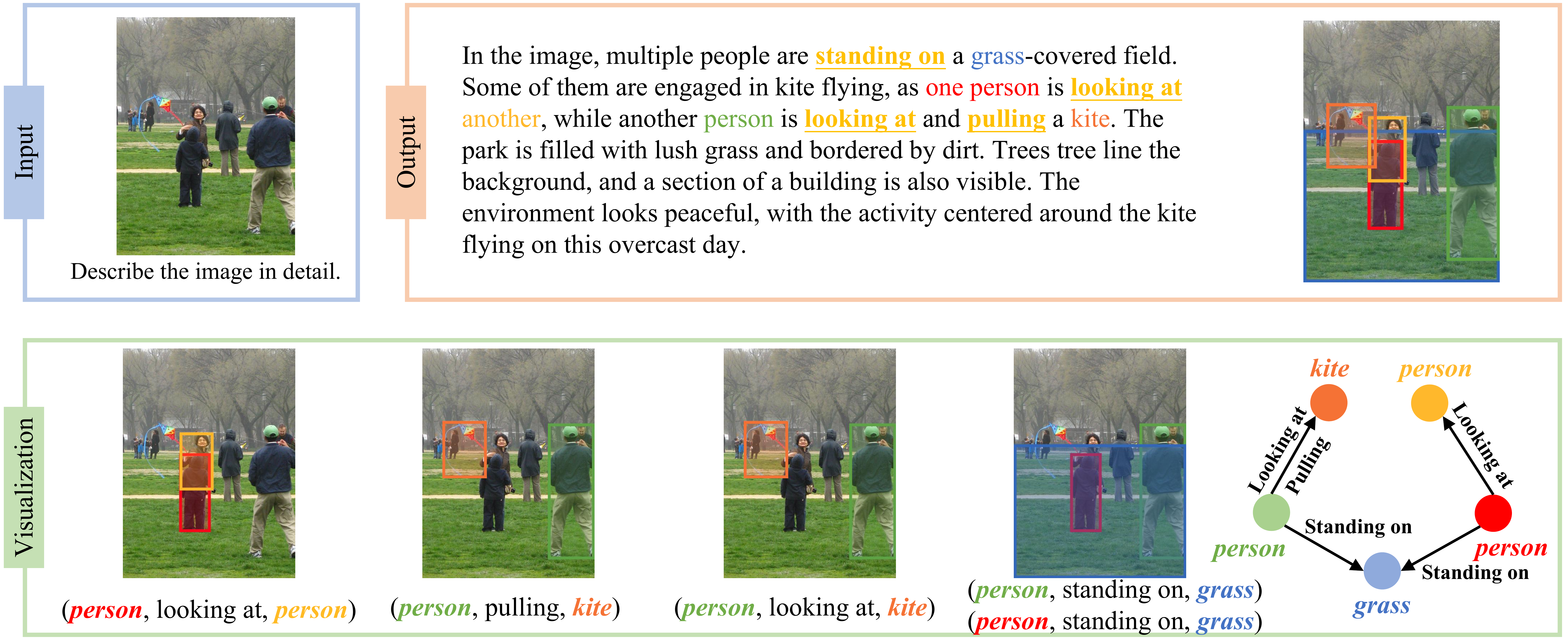}
     \caption{}
\end{subfigure}
\caption{
    \textbf{Data Examples of Detailed Description task in {\datasetname}}.
}
\label{fig:appendix-as-v2-global}
\end{figure}

\begin{figure}[h!]
\hsize=\linewidth
\centering
\begin{subfigure}{\linewidth}
    \centering
    \includegraphics[width=\linewidth]{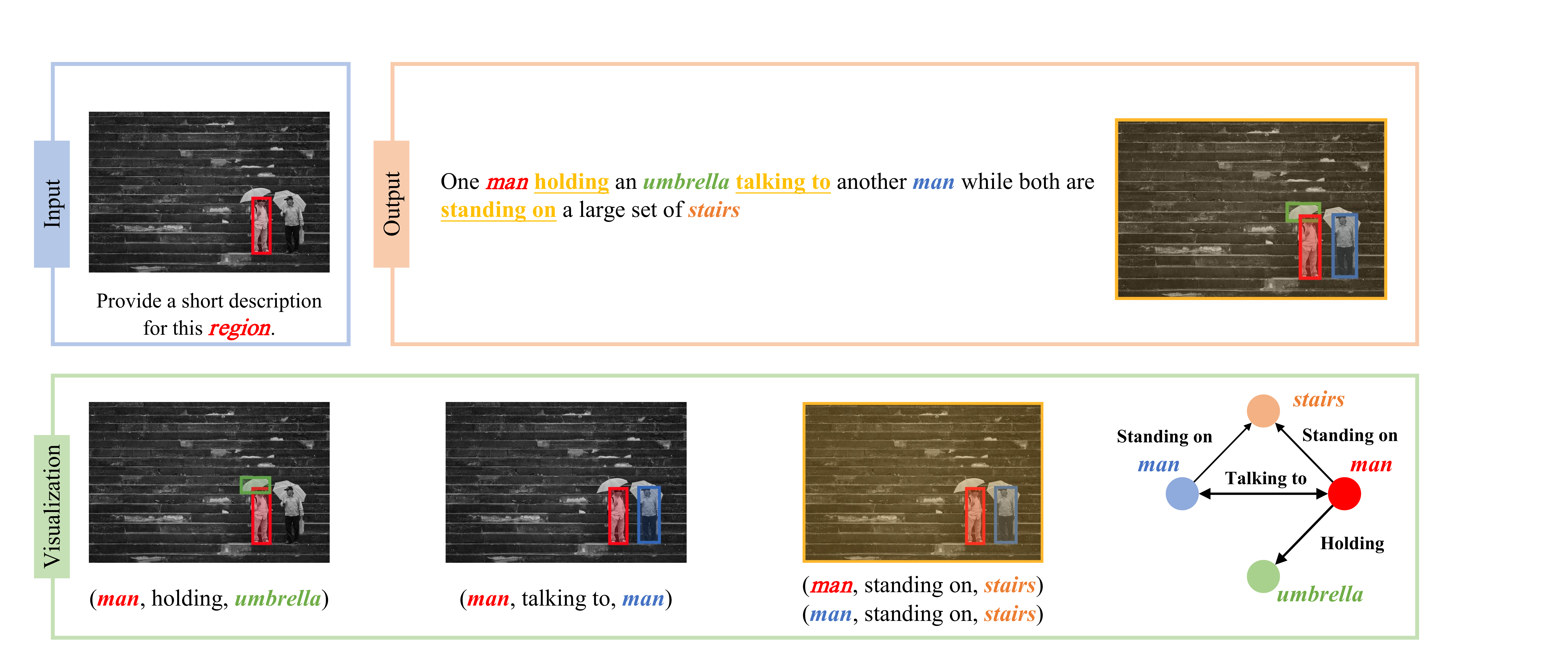}
    \caption{}
\end{subfigure}    
\begin{subfigure}{\linewidth}
     \centering
     \includegraphics[width=\linewidth]{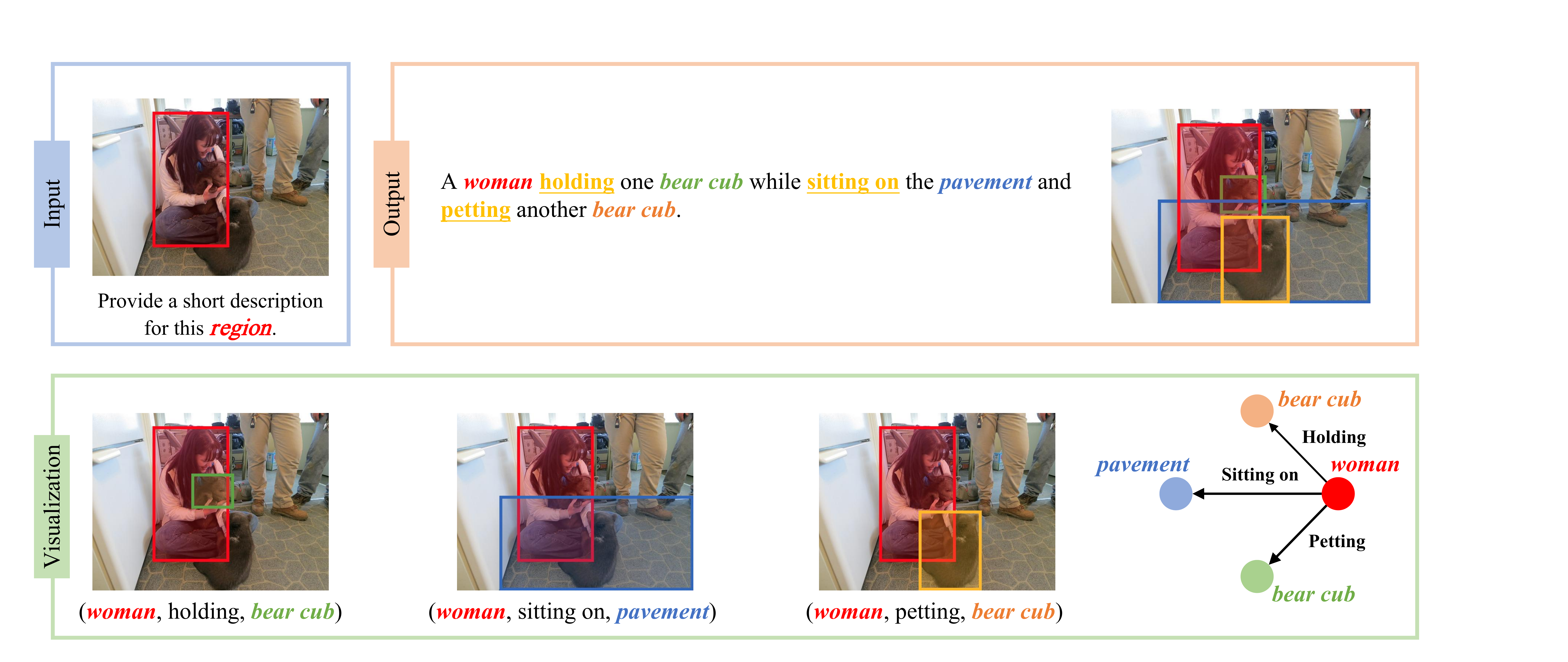}
     \caption{}
\end{subfigure}
\begin{subfigure}{\linewidth}
     \centering
     \includegraphics[width=\linewidth]{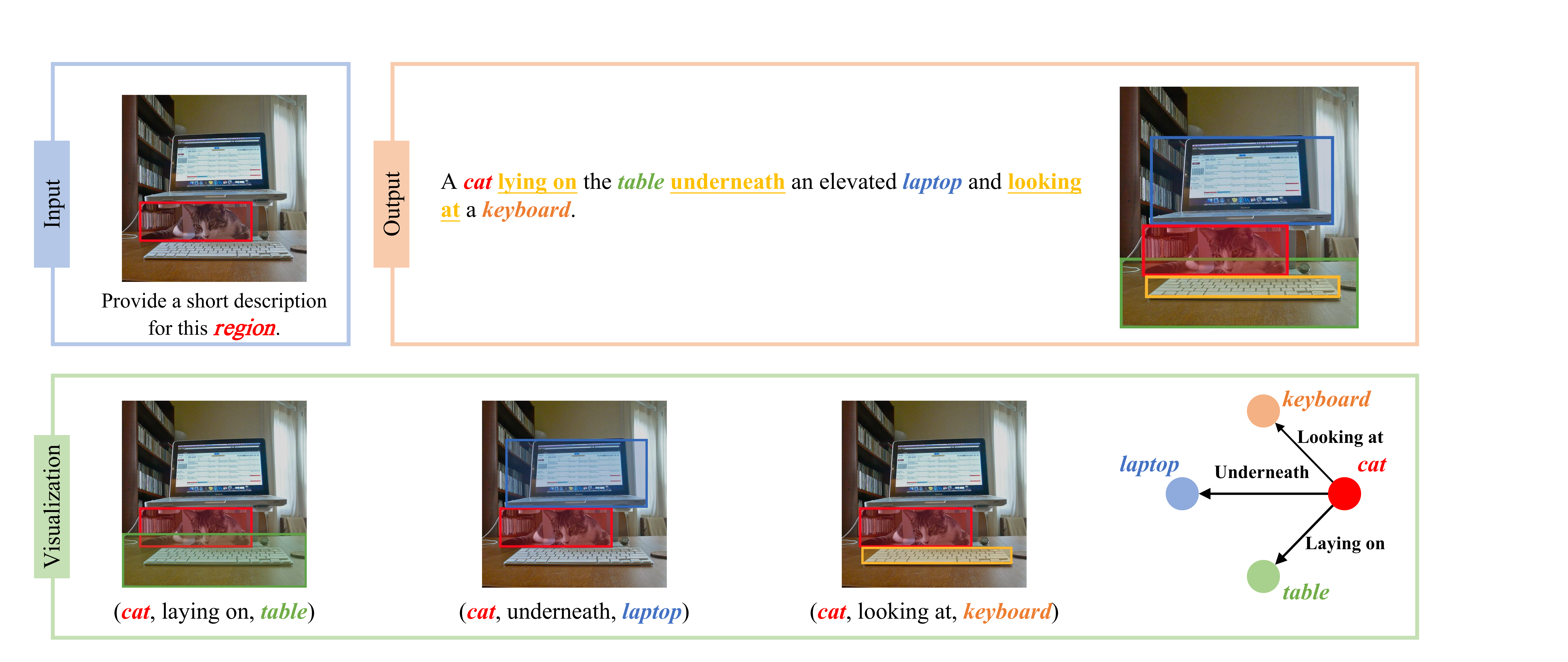}
     \caption{}
\end{subfigure}
\caption{
    \textbf{Data Examples of Region Captioning task in {\datasetname}}.
}
\label{fig:appendix-as-v2-region}
\end{figure}

\begin{figure}[h!]
\hsize=\linewidth
\centering
\begin{subfigure}{\linewidth}
    \centering
    \includegraphics[width=\linewidth]{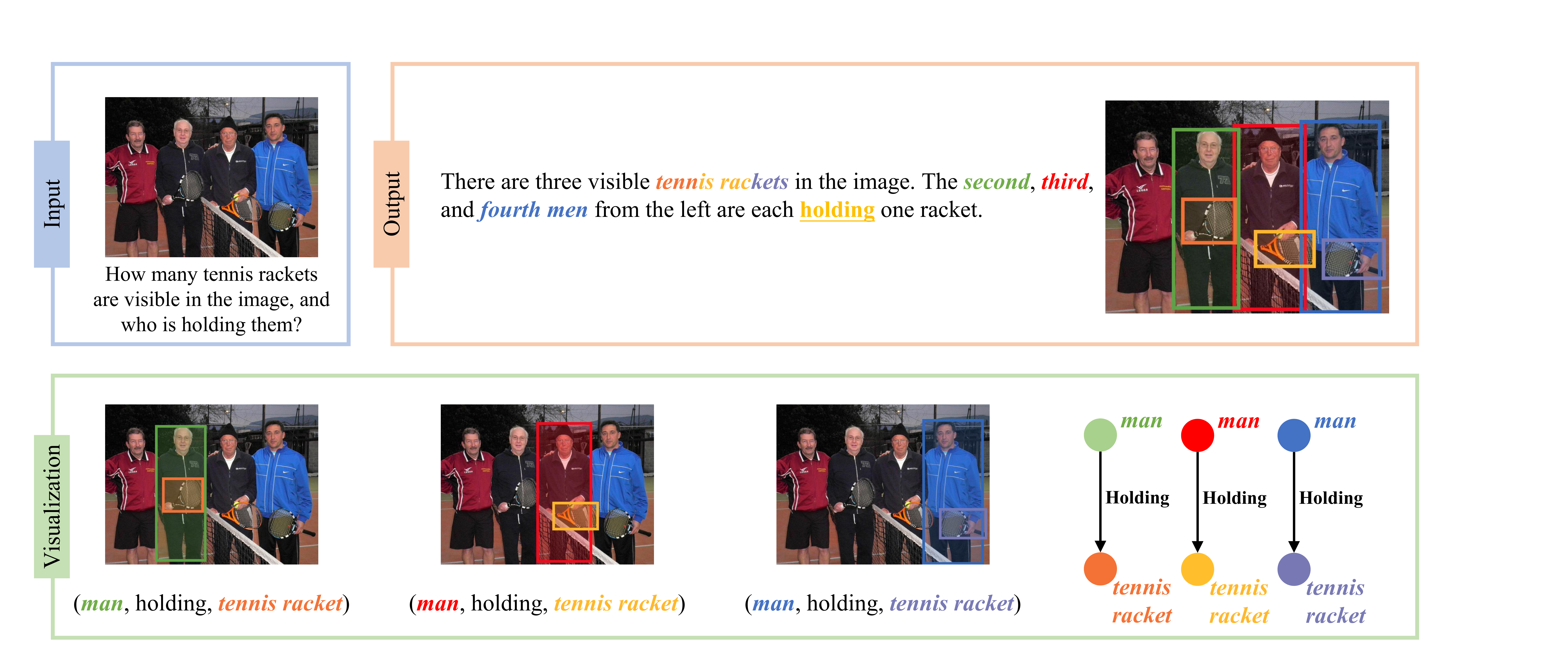}
    \caption{}
\end{subfigure}    
\begin{subfigure}{\linewidth}
     \centering
     \includegraphics[width=\linewidth]{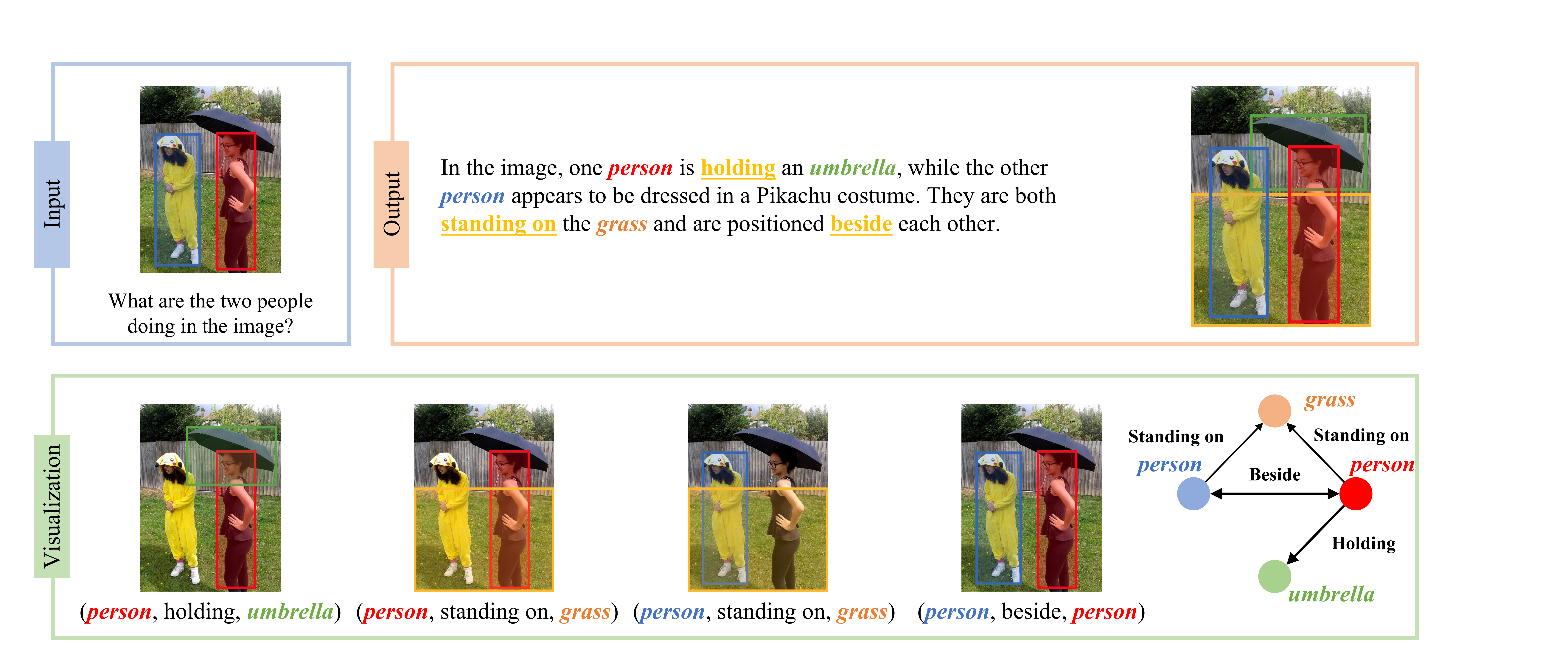}
     \caption{}
\end{subfigure}
\begin{subfigure}{\linewidth}
     \centering
     \includegraphics[width=\linewidth]{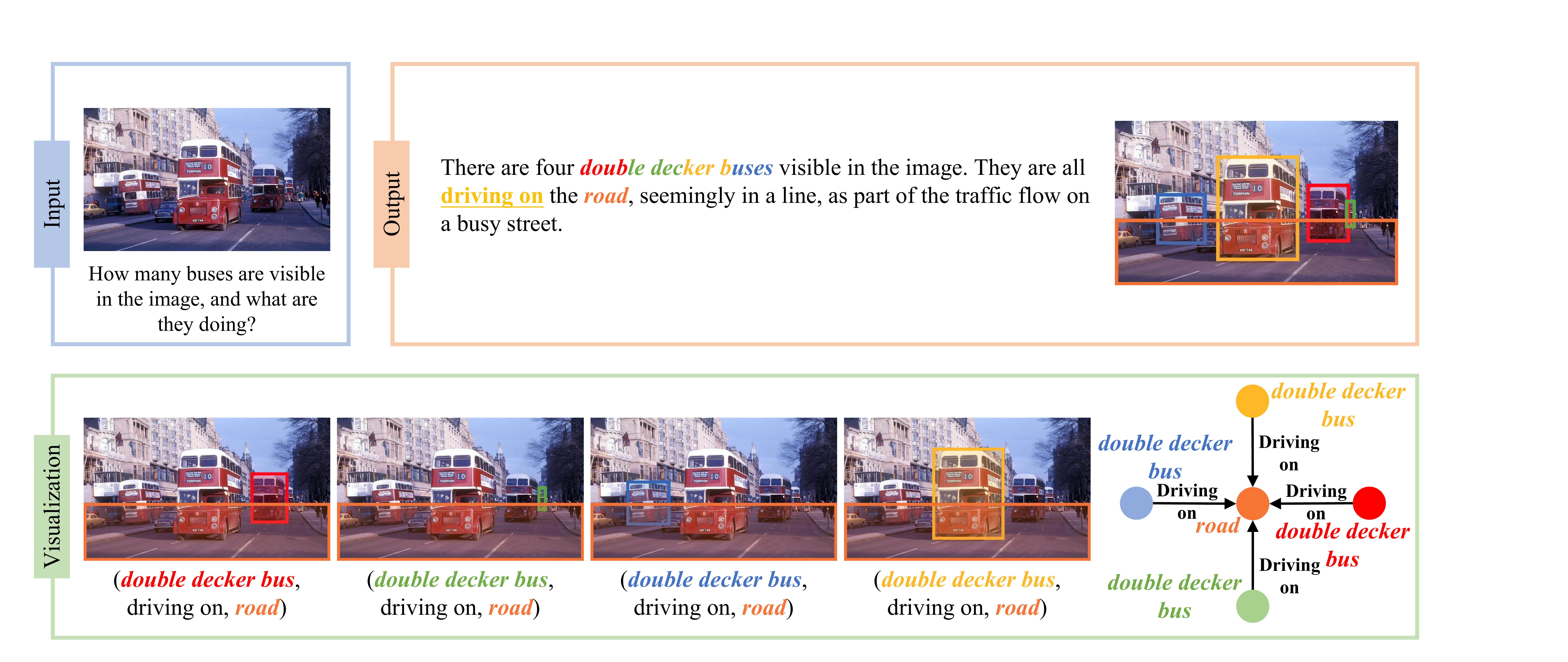}
     \caption{}
\end{subfigure}
\caption{
    \textbf{Data Examples of Conversation task in {\datasetname}}.
    Due the space limitations, we exhibit only one turn for each conversation.
}
\label{fig:appendix-as-v2-conv}
\end{figure}

More data examples of {\datasetname} are shown in \cref{fig:appendix-as-v2-global,fig:appendix-as-v2-region,fig:appendix-as-v2-conv}.
Besides, prompts used to generate detailed description data are shown in \cref{tab:prompt_task_desc,tab:prompt_full_example}.

\begin{table*}[h!]

\footnotesize
\centering
\caption{
For each query,
the system info explains the \textbf{task description} and the in-context-learning examples are presented in the form of multi-turn conversation.
For each turn, the input \VarSty{query[`context']} consists of
(1) the {\textbf{image}} to be annotated,
(2) the {\textbf{captions annotations}} of this image,
(3) the {\textbf{location annotations}},
as well as (4) the {\textbf{relation annotations}}.
The output \VarSty{query[`context']} comprises the manually annotated scene graph conversation data.
In this example, we provide the task description for the Detailed Description data in the relation conversation.
}
\label{tab:prompt_task_desc}
    
\begin{minipage}{\linewidth}
\begin{tcolorbox} 
\begin{tabular}{p{\linewidth}}
\begin{minipage}{\linewidth}
\VarSty{messages} = [\{\var{"role":"system", "content":} f"""You are an AI visual assistant that can analyze a single image. You receive one image and five sentences, each describing this image you are observing. In addition, specific object locations within the image are given, along with detailed coordinates. These coordinates are in the form of bounding boxes, represented as \texttt{[x1, y1, x2, y2]} with int numbers ranging from 0 to 999. These values correspond to the top left x, top left y, bottom right x, and bottom right y. Note that these coordinates are normalized. Besides, the scene graph of this image is also provided as a list of tuples. Each tuple is represented as (subject, bounding box of the subject, object, bounding box of the object, predicate).
\\ \\
Using the provided caption, bounding box, and scene graph information, describe the scene in a detailed manner. If there are errors in the caption, please ignore them and do not point them out in your description.
\\ \\
Instead of directly mentioning the bounding box coordinates, utilize this data to explain the scene using natural language with its bounding box in the format like "\texttt{<ref>object</ref><box>[[x1, y1, x2, y2]]</box>}". When mentioning the predicate between two objects, you should mention it in the format like "\texttt{<pred>predicate</pred><box>[[x1, y1, x2, y2]]</box><box>[[x3, y3, x4, y4]]</box>}", where "\texttt{<box>[[x1, y1, x2, y2]]</box>}" denotes the bounding box coordinates of the subject and "\texttt{<box>[[x3, y3, x4, y4]]</box>}" denotes the bounding box coordinates of the object. Include details like object counts, position of the objects, relative position between the objects.
\\ \\
When using the information from the caption, coordinates, or scene graph, directly explain the scene, and do not mention that the information source is the caption or the bounding box or the scene graph.  You should mention all tuples and predicates included in the scene graph in the generated caption. Make sure that the box following the \texttt{<pred>predicate</pred>} has already been mentioned after a \texttt{<ref>object</ref>}."""\}\\
]

\For{\VarSty{sample} \bf in \VarSty{fewshot\_samples}}{
\var{\VarSty{messages}.append(\{"role":"user","content":\VarSty{sample[`context']}\})}\;
\var{\VarSty{messages}.append(\{"role":"assistant","content":\VarSty{sample[`response']}\})}\;
}
\var{\VarSty{messages}.append(\{"role":"user","content":`\textbackslash n'.join(\VarSty{query})\})}
\end{minipage}
\end{tabular}
\end{tcolorbox}
\end{minipage}
\end{table*}

\begin{table*}[h!]
\caption{
One example to illustrate the instruction-following data. The top block shows the contexts such as captions, locations, relations and images used to prompt GPT, and the bottom block shows the three types of responses.
Note that the visual image is also used to prompt GPT.
}
\label{tab:prompt_full_example}

\centering
\footnotesize

\begin{minipage}{1.0\columnwidth}
\begin{tcolorbox}
\begin{tabular}{p{0.97\columnwidth} c}
\VarSty{ {\bf Context type 1: Captions} } &\\
A pretty young lady holding a dark colored umbrella.& \\
A girl in a Pikachu suit standing beside a girl holding an umbrella. & \\
A woman is wearing a raincoat and another woman is holding an umbrella over her head. & \\
two girls stand in a yard, one holds an umbrella and the other is dressed as pikachu.
\hspace{6.2cm}
\multirow{5}{*}{\includegraphics[height=3.0cm]{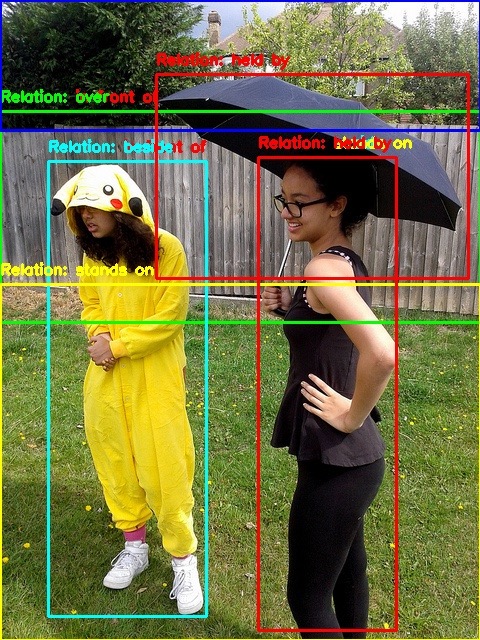}}& \\
Two women who are standing in the grass. & \\

\VarSty{ {\bf Context type 2: Locations} } & \\
person: [101, 252, 430, 963]& \\
person: [539, 246, 826, 984]& \\
grass: [0, 444, 999, 999]& \\
$\cdots$& \\

\VarSty{ {\bf Context type 3: Relations} } & \\
(person, [101, 252, 430, 963], grass, [0, 444, 999, 999], standing on) & \\
(person, [101, 252, 430, 963], person, [539, 246, 826, 984], beside)& \\
(person, [539, 246, 826, 984], grass, [0, 444, 999, 999], standing on) & \\
$\cdots$& \\

\hrulefill & \\

\VarSty{ {\bf Response type 1: Detailed Description} } & \\
In the image, two <ref>people</ref><box>[[101, 252, 430, 963], [539, 246, 826, 984]]</box> are <pred>standing on</pred><box>[[101, 252, 430, 963], [539, 246, 826, 984]]</box><box>[[0, 444, 999, 999]]</box> a <ref>grass</ref><box>[[0, 444, 999, 999]]</box>. 
$\cdots$ & \\

\VarSty{ {\bf Response type 2: Region Captioning} }& \\
person: [539, 246, 826, 984]& \\
A <ref>\-girl\-</ref>\-<box>\-[[101, 252, 430, 963]]\-</box> in a Pikachu suit standing <pred>\-beside\-</pred>\-<box>\-[[101, 252, 430, 963]]\-</box>\-<box>\-[[539, 246, 826, 984]]\-</box> a <ref>\-girl\-</ref>\-<box>\-[[539, 246, 826, 984]]\-</box>.&\\
$\cdots$

\VarSty{ {\bf Response type 3: Conversation} }& \\
Question: \\
What are the two <ref>people</ref><box>[[101, 252, 430, 963], [539, 246, 826, 984]]</box> doing in the image? \\
--- \\
Answer: \\
In the image, one <ref>person</ref><box>[[539, 246, 826, 984]]</box> is <pred>holding</pred><box>[[539, 246, 826, 984]]</box><box>[[326, 117, 976, 435]]</box> an <ref>umbrella</ref><box>[[326, 117, 976, 435]]</box>,
$\cdots$
\\
$\cdots$
\end{tabular}
\end{tcolorbox}
\end{minipage}
\end{table*}

\clearpage

\section{The All-Seeing Model v2}
\label{sec:appendix_asmv2}

\subsection{Implementation Details}
\label{sec:method-details}

\begin{table}[t]
\renewcommand{\arraystretch}{0.9}
\setlength{\belowcaptionskip}{1.5mm}
\footnotesize
\centering
\caption{
    \textbf{Details of the instruction-tuning data for {\modelname} in stage 2}.
    We collect a wide range of high-quality data, totaling approximately 4 million samples.
}
\label{tab:stage2_sft_data}

\begin{tabular}{l|c|l}
    \textbf{Task} & \textbf{\#Samples} & \textbf{Dataset} \\
    \hline
    
    \rowcolor{mygray}
    Captioning & 124K  & TextCaps~\cite{sidorov2020textcaps}, ShareGPT4V~\cite{chen2023sharegpt4v}\\

    & & VQAv2~\cite{goyal2017vqav2}, GQA~\cite{hudson2019gqa}, OKVQA~\cite{marino2019ok}, A-OKVQA~\cite{schwenk2022aokvqa}, \\
    \multirow{-2}{*}{VQA} & \multirow{-2}{*}{314K} & ScienceQA~\cite{lu2022sqa}, CLEVR~\cite{johnson2017clevr}, Visual7W~\cite{zhu2016visual7w} \\

    \rowcolor{mygray}
    OCR & 157K & ST-VQA~\cite{biten2019stvqa}, LLaVAR~\cite{zhang2023llavar}, OCR-VQA~\cite{mishra2019ocrvqa}, DocVQA~\cite{clark2017docqa} \\
    
    Grounding & 643K & RefCOCO/+/g~\cite{kazemzadeh2014refcoco,mao2016refcoco_plus_g} \\
    
    \rowcolor{mygray}
    RegionVQA & 2.3M & RefCOCOg~\cite{mao2016refcoco_plus_g}, VG~\cite{krishna2017vg}, VCR~\cite{zellers2019vcr}, AS-Core~\cite{wang2023allseeing} \\

    Conversation & 500K & LLaVA-Instruct~\cite{liu2023llava}, SVIT~\cite{zhao2023svit}, LRV~\cite{liu2023lrv}, AS-V2 (ours) \\
    
    \rowcolor{mygray}
    Text & 40K & ShareGPT~\cite{sharegpt} \\

\end{tabular}
\end{table}

\subsubsection{Training Stage 1.}
The global batch size is set to 256 in the pre-training phase and 128 in the instruction-tuning phase. We employ the AdamW optimizer~\cite{adamw} with the $\beta_1$ of 0.9, the $\beta_2$ of 0.999, and the weight decay of $0$. The learning rate is initialized as $1\times10^{-3}$ for the pre-training phase and $2\times10^{-5}$ for the instruction-tuning phase. Both phases include a linear warmup that lasts until the first 3\% of training steps.
The warmup is followed by a cosine decay strategy with a minimum learning rate of 0.
We only train the vision-language connector in the pre-training phase while both the vision-language connector and the language model are trainable in the instruction-tuning phase.
We train the model for 1 epoch in both phases.
The image resolution of {\modelname} is set to 336 $\times$ 336.

\subsubsection{Training Stage 2.}
The global batch size is set to 512 and the learning rate is initialized as $2\times10^{-5}$ in both the pre-training phase and the instruction-tuning phase.
The language model and vision-language connector are trainable in both phases while the vision encoder is always frozen.
We train the model for 5000 steps in the pre-training phase and 1 epoch in the instruction-tuning phase.
The other settings remain the same as the instruction-tuning phase of Stage 1.

\subsection{Predicate Classification}
\label{sec:exp-predcls}
\begin{table}[t]
\centering
\caption{
\textbf{Recall scores on Predicate Classification task}.
}
\label{tab:pred_cls}
\begin{tabular}{@{}lcccccc@{}}
\toprule
\multirow{2}{*}{Method} & \multicolumn{6}{c}{Predicate Classification}                                         \\ \cmidrule(l){2-7} 
                        & R@20          & mR@20         & R@50 & mR@50         & R@100         & mR@100        \\ \midrule
IMP~\cite{xu2017IMP}                     & 30.5          & 9.0           & 35.9 & 10.5          & 38.3          & 11.3          \\
MOTIFS~\cite{zellers2018MOTIFS}                  & 45.1          & 19.9          & 50.5 & 21.5          & 52.5          & 22.2          \\
VCTree~\cite{tang2019vctree}                  & \textbf{45.9} & \textbf{21.4}          & \textbf{51.2} & 23.1          & \textbf{53.1}          & 23.8          \\
GPSNet~\cite{lin2020gpsnet}                  & 38.8          & 17.1          & 46.6 & 20.2          & 50.0          & 21.3          \\
\rowcolor{mygray}
ASMv2 (ours)                   & 17.6          & \textbf{21.4} & 25.9 & \textbf{34.4} & 32.6 & \textbf{44.5} \\ \bottomrule
\end{tabular}
\end{table}

\begin{figure}[t]
\hsize=\textwidth
\setlength{\abovecaptionskip}{1.5mm}
\centering
\begin{subfigure}{0.45\textwidth}
    \centering
    \includegraphics[width=0.8\textwidth]{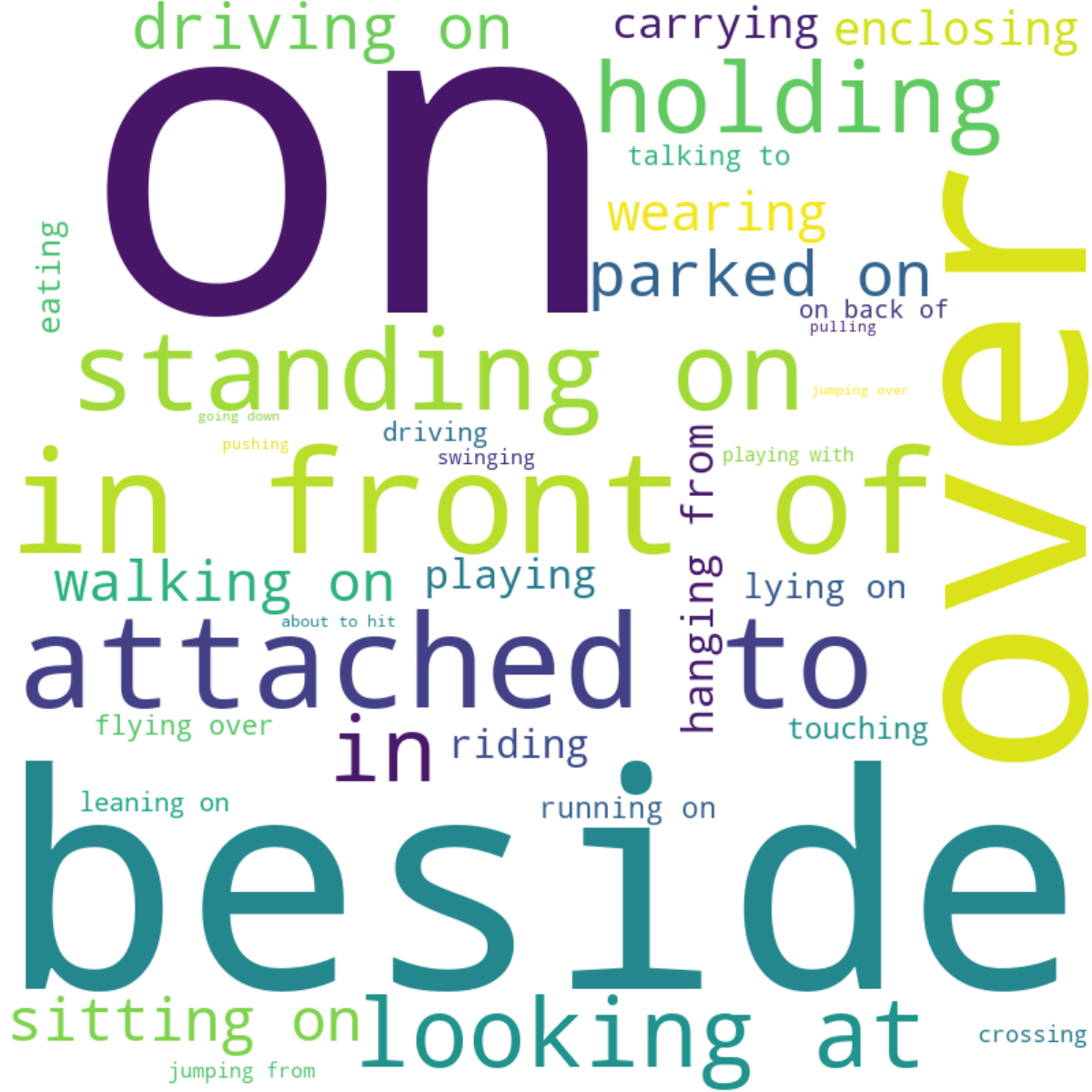}
    \caption{}
    \label{fig:wc-a}
\end{subfigure}    
\begin{subfigure}{0.45\textwidth}
     \centering
     \includegraphics[width=0.8\textwidth]{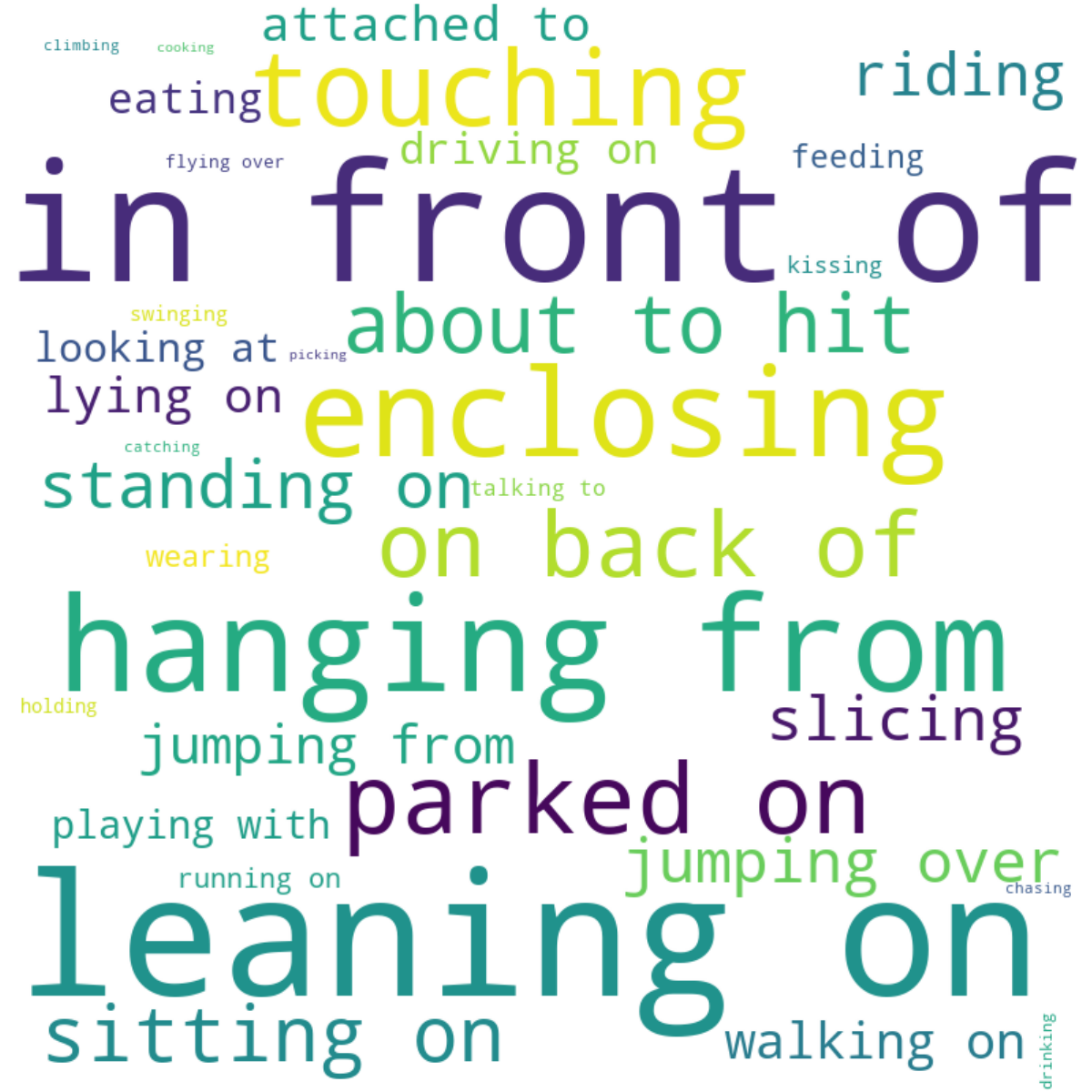}
     \caption{}
     \label{fig:wc-b}
\end{subfigure}
\caption{
    \textbf{Word Clouds for evaluation data in PSG.}
    \cref{fig:wc-a} visualizes the distribution of ground-truth predicates while \cref{fig:wc-b} visualizes those predicted by {\modelname}.
}
\label{fig:word_cloud_pred_cls}
\end{figure}

In this section, we evaluate the relation comprehension capability of our model through the Predicate Classification task (PredCls) on the Panoptic Scene Graph (PSG) dataset~\cite{yang2022psg}.
Compared to the Open-ended Scene Graph Generation task, PredCls aims to generate a scene graph given the ground-truth object labels and localization, focusing on the relation prediction performance without the interference of the detection performance.

\subsubsection{Evaluation Setup.}

Assuming that the number of ground-truth objects is N, we query the model for $N\times\left(N-1\right)$ times, considering each of the ground-truth objects as the subject or object.
For each query, we ask the model ``\textit{What is the relation between the \texttt{<subject>} and the \texttt{<object>}? Answer the question using a single word or phrase.}'' and employ a vocabulary ranking method~\cite{brown2020gpt3} to generate the scores for each predicate label.
Following prior works~\cite{yang2022psg,tang2019vctree}, we report the Recall and mean Recall (mRecall) here.

\subsubsection{Results.}

As shown in \cref{tab:pred_cls}, our {\modelname} demonstrates competitive performance on the Predicate Classification task within the PSG dataset.
Specifically, our {\modelname} achieves superior performance in mRecall but is inferior in Recall.
For instance, our {\modelname} significantly outperforms VCTree by 11.3 points in mR@50 and 20.7 points in mR@100, while it falls behind in terms of Recall.
These results stem from the PSG dataset's inherently imbalanced distribution of predicate labels, where broad predicates such as ``on'' and ``over'' are more frequent. 
As depicted in \cref{fig:word_cloud_pred_cls}, our {\modelname} is less likely to predict these common but general predicates. Instead, it tends to predict more specific and less frequent predicates, like ``standing on'' and ``parked on'', resulting in superior mRecall while inferior Recall.
These results underline our model's deeper and more detailed comprehension of visual relations.

\subsection{Ablation Study}
\begin{table}[t]
\centering
\caption{
\textbf{Ablation results}.
We report the $Q \rightarrow AR$ accuracy for VCR~\cite{zellers2019vcr} and the overall accuracy for CRPE.
REC denotes the average accuracy score across RefCOCO~\cite{kazemzadeh2014refcoco}, RefCOCO+~\cite{mao2016refcoco_plus_g}, and RefCOCOg~\cite{mao2016refcoco_plus_g}.
}
\label{tab:ablation}

\begin{tabular}{@{}lccccc@{}}
\toprule
Ablation Settings                   & MME    & MM-Vet & REC & VCR  & CRPE \\ \midrule
ASMv2                   & 1621.0 & 41.3   & 87.4           & 78.4 & 64.5           \\ \midrule
- Stage1 Training       & 1527.6 & 35.6   & 86.1           & 78.7 & 64.7           \\
- Relation Conversation & 1554.6 & 42.2   & 86.6           & 77.0 & 55.3           \\ \bottomrule
\end{tabular}
\end{table}

In this section, we ablate the training settings of {\modelname}. The experimental settings are the same as those discussed in \cref{sec:method-details}.
As shown in \cref{tab:ablation}, the two-stage training process of {\modelname} and the utilization of relation conversation data is essential for achieving excellent performance on both image-level and region-level benchmarks simultaneously.
We can observe that skipping the first training stage leads to significant performance degradation on MME~\cite{fu2023mme}, MM-Vet~\cite{yu2023mmvet}, and Referring Expression Comprehension (REC) Benchmarks~\cite{kazemzadeh2014refcoco,mao2016refcoco_plus_g}, indicating that the model struggles to understand visual information at both the image and region levels simultaneously without the two-stage training strategy.

Furthermore, removing relation conversation data from the training corpora results in inferior performance on REC~\cite{kazemzadeh2014refcoco,mao2016refcoco_plus_g}, VCR~\cite{zellers2019vcr}, and CRPE.
The performance on MME~\cite{fu2023mme} also experiences a drop of about 66.4 points, primarily due to a decrease in the count score, from 170.0 to 155.0, and a decrease in the position score, from 163.3 to 133.3.
These results demonstrate the effectiveness of our relation conversation data in improving capabilities for region-level visual information understanding and relation comprehension.

\clearpage

\section{The Circular-based Relationship Probing Evaluation}
\label{sec:appendix_crpe}

\begin{figure}[!h]
\centering
\includegraphics[width=\linewidth]{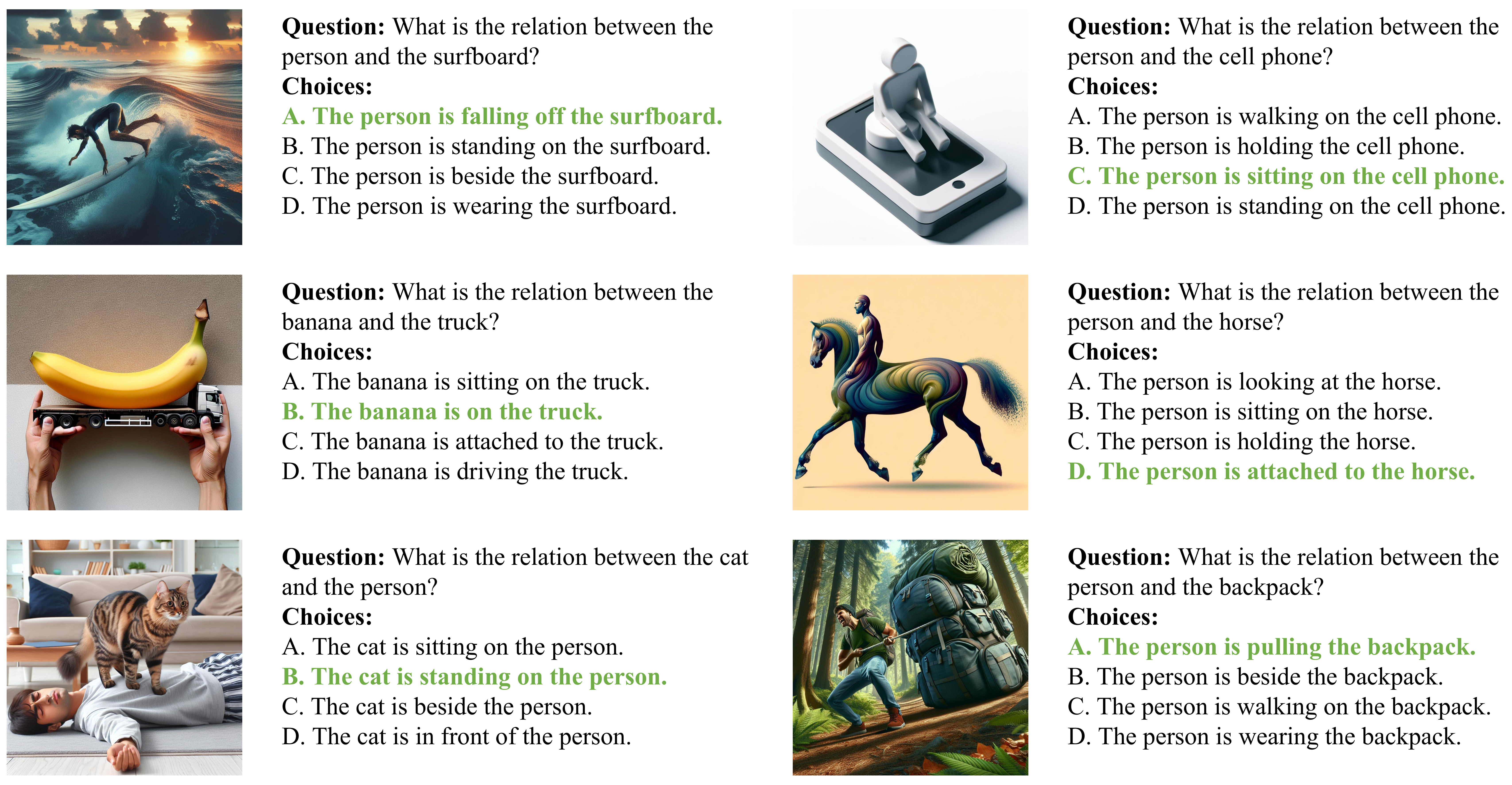}
\caption{
    \textbf{Data examples of abnormal data in the {\benchmarkname}}.
}
\label{fig:appendix-crpe-abnormal}
\end{figure}

In this section, we present more examples of abnormal data in {\benchmarkname} in \cref{fig:appendix-crpe-abnormal}.

\end{document}